\documentclass[a4paper,fleqn]{cas-sc}

\usepackage[numbers]{natbib}

\usepackage{multirow}
\usepackage{amsmath}
\usepackage{amssymb}
\usepackage{graphicx}
\usepackage{color}
\usepackage[switch]{lineno}
\usepackage{lipsum}
\usepackage{xcolor}
\usepackage{relsize}
\usepackage{float}
\usepackage{color, colortbl}
\definecolor{Gray}{RGB}{
231, 245, 255}
\newcolumntype{a}{>{\columncolor{Gray}}c}
\usepackage{subcaption}
\usepackage{svg}
\newcommand{\orcid}[1]{\href{https://orcid.org/#1}{\includegraphics[width=10pt]{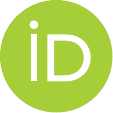}}}
\newcommand{\grayline}{\arrayrulecolor{gray}\hline\arrayrulecolor{black}}

\DeclareMathOperator*{\argmax}{arg\,max}

\usepackage{arydshln}

\usepackage{pifont}
\newcommand{\cmark}{\ding{51}}%
\newcommand{\xmark}{\ding{55}}%

\usepackage{textcomp,array,booktabs}
\usepackage{makecell}

\usepackage{adjustbox}

\def\tsc#1{\csdef{#1}{\textsc{\lowercase{#1}}\xspace}}
\tsc{WGM}
\tsc{QE}
\tsc{EP}
\tsc{PMS}
\tsc{BEC}
\tsc{DE}

\begin{document}
\let\WriteBookmarks\relax
\def\floatpagepagefraction{1}
\def\textpagefraction{.001}
\shorttitle{Dynamic self-ensemble on multi-exit residual ViT for small-data image classification}

\title [mode = title]{Dynamic self-ensemble on multi-exit residual vision transformer for small-data image classification}                      


\author[1]{Stephen Ekaputra Limantoro}[type=editor]
\ead{stephenekaputra@gmail.com}

\affiliation[1]
{
    organization={Department of Electrical Engineering and Computer Science, National Yang Ming Chiao Tung University},
    city={Hsinchu},
    postcode={30010},
    country={Taiwan}
}

\credit{Writing - original draft, Software, Methodology, Conceptualization}

\author[1]{Yan-Kai Wang}[style=chinese]
\ead{ykwang.ee09@nycu.edu.tw}

\credit{Investigation, Validation, Software}

\author[2]{Ching-Chun Huang}[style=chinese]
\ead{chingchun@nycu.edu.tw}

\credit{Supervision, Writing - review \& editing, Formal analysis}

\affiliation[2]
{
    organization={Department of Computer Science, National Yang Ming Chiao Tung University},
    city={Hsinchu},
    postcode={30010},
    country={Taiwan}
}

\author[1]{Hong-Han Shuai}[style=chinese]
\ead{hhshuai@nycu.edu.tw}

\credit{Supervision, Writing - review \& editing, Validation}

\author[3,4]{Wen-Huang Cheng}[style=chinese]
\cormark[1]
\ead{wenhuang@csie.ntu.edu.tw}

\credit{Supervision, Writing - review \& editing, Validation, Formal analysis, Resources}

\affiliation[3]
{
    organization={Department of Computer Science and Information Engineering, National Taiwan University},
    city={Taipei},
    postcode={10617},
    country={Taiwan}
}

\affiliation[4]
{
    organization={College of Engineering and Computer Science, VinUniversity},
    city={Hanoi},
    country={Vietnam}
}

\author[5]{Jungwoo Huh}[style=chinese]
\ead{gjwjddn9@yonsei.ac.kr}

\credit{Writing - review \& editing, Visualization, Validation}

\affiliation[5]
{
    organization={Department of Electrical and Electronic Engineering, Yonsei University},
    city={Seoul},
    postcode={03722},
    country={South Korea}
}

\author[5]{Sanghoon Lee}[style=chinese]
\ead{slee@yonsei.ac.kr}

\credit{Supervision, Validation, Resources}

\cortext[cor1]{Corresponding author}

\begin{abstract}
Vision Transformer (ViT) has garnered increased attention for image classification in recent years. However, ViT is notorious for its data-hungry problem as it achieves high-performance results when only trained on large-size datasets. In this paper, we study how ViT specifically works on small-scale datasets with limited data where the availability of image data is often constrained in practice, especially in fields such as medical imaging. To mitigate the issue, we designed a multi-stage residual ViT, dubbed ReSEViT, with the self-ensemble method to utilize multi-level information from intermediate features. Due to the complex decision boundaries for classifications, relying on a single classifier can lead to prediction errors. Hence, we propose a new learning method for ViT, named self-ensemble learning, which ensembles multi-stage classifiers in a single model. Our self-ensemble comprises a novel approach, Dynamic Learner Selection (DLS), to adaptively select the importance of different stages. ReSEViT incorporates residual token embeddings which is a ResNet-like block to model effective local structures alongside with transformer blocks in each stage. The integration of residual tokens and self-ensemble
learning into a multi-stage ViT highlights the reduction in redundancy, prediction error and variance while maintaining an efficient model size for small-scale image datasets.
\end{abstract}

\begin{keywords}
Vision transformer 
\sep 
Self-ensemble learning
\sep
Efficient model
\sep
Small-data image classification
\end{keywords}

\maketitle

\section{Introduction}

Recent advancements in deep learning models, particularly Transformers \cite{transformer}, have significantly impacted various vision tasks \textit{i.e.}, image classification \cite{vit,cvt,pvt}, object detection \cite{deformdetr}, and image processing \cite{imgprocess}. Among these, image classification is a fundamental building block that supports the development of more complex applications \cite{fasterrcnn,maskrcnn}. ViT \cite{vit} introduced the first fully transformer-based model that attains competitive results on large-scale image classification benchmarks. Despite these advances, the performance of ViT is still inferior to that of CNN-based models when trained from scratch on medium-sized datasets, \textit{e.g.}, ImageNet-1K. This disparity arises as the transformers lack the locality inductive biases (local receptive field, translation invariance, and spatial subsampling \cite{localbias}), which are inherent in CNNs. Consequently, training ViT from scratch with limited data yields suboptimal performance, while its high computational cost further challenges deployment on resource-constrained edge devices, IoT devices, consumer electronics, and wearable medical devices that rely on lightweight models \cite{sxsformer, pdformer, pest_iot, sqvit, household_transformer, medmnistv1, smalldatasetaugmentation}.

\begin{figure}[pos=ht]
\centering
\includegraphics[scale=0.5]{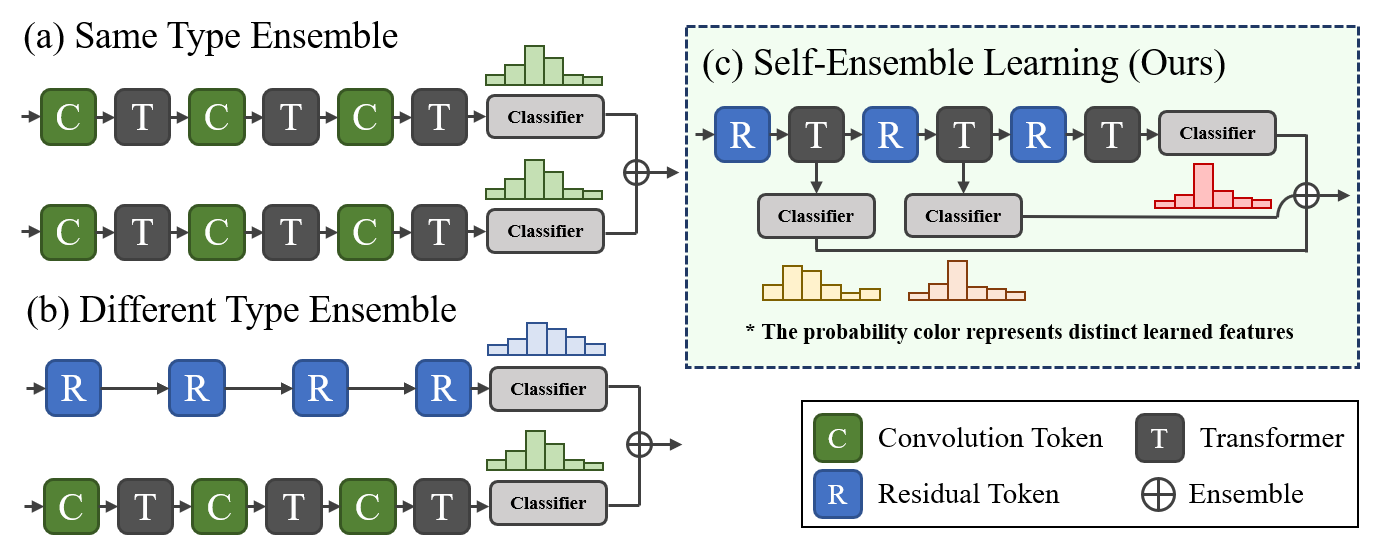}
\caption{Illustration of traditional ensemble and our self-ensemble learning for our ReSEViT. The traditional ensemble methods integrating multiple models (\textit{e.g.} (a) two CvTs, (b) ResNet and CvT) separately are computationally heavier and only suggest high-level representation producing high-similar semantic outputs. (c)  The prediction of ReSEViT is more accurate as multi-level representation evades the risk of semantic redundancy where the transformer in each stage inherits stronger local inductive biases from the residual token. In addition, both training and testing time could be drastically reduced with the self-ensemble learning.}
\label{fig:problemsetting}
\end{figure}

Several works \cite{nest,cct,howvitsmall,efficientvitsmall} have introduced strategies for training ViT from scratch on small datasets with either model architecture modifications or self-supervised techniques to solve the data-hungry problem. It takes place when ViT underperforms due to a limited amount of data. A recent work, Compact Convolutional Transformer (CCT) \cite{cct}, introduced a simple CNN-Transformer hybrid approach that can be well-trained from scratch on small-scale datasets (\textit{e.g.}, CIFAR) by employing convolutional tokenization and a transformer encoder with sequence pooling. This architecture design aims to take advantage of the inductive biases from CNNs and integrate them with a transformer. To further improve the ViT model, one straightforward strategy is to use the existing ensemble methods~\cite{bagging,adaboost,ensemblesketchrecognition}. However, ensemble methods typically lead to a significant increase in the number of parameters, which is often impractical for real-time applications, especially for large models. For example, CCT-14/7$\times$2 is a single-stage type model requiring 22.36M parameters for the ImageNet dataset~\cite{cct}, which may expand to more than 100M if we ensemble five models with similar sizes in a parallel way. Therefore, we proposed ReSEViT, a residual vision transformer with the self-ensemble method. The self-ensemble method is a novel ensemble method using only one deep learning model for an image classification task. Such a hierarchical structure ensemble strategy has been explored previously on object detection tasks, \textit{i.e.}, FPN \cite{fpn}, but research on image classification has barely been touched. 

In our approach, we first design a ViT backbone with a multi-stage structure to enable self-ensemble learning. Specifically, inspired by CvT~\cite{cvt}, it adopts a three-stage model consisting of convolutional token embedding and a convoluted transformer block in each stage. In our method, a residual token embedding \cite{resnet} is adopted to model effective local structures in each stage. Notably, the ResNet block is not a new architecture. However, it is employed to provide an efficient strategy for ViT's data hunger on small image datasets. Similar to the works of \cite{mobilevit, shiftvit, poolformer}, parts of the architecture are replaced by existing methods, such as the MobileNet block, shift operation, and pooling layer, to demonstrate the effect on the model for a specific purpose. We hypothesize that the residual token enables our self-ensemble to perform better and more efficiently for small-scale datasets since the local inductive biases for the transformer are denser in each stage. Intuitively, stronger feature representations at each stage lead to a compounding effect in the self-ensemble method resulting in high performance while preserving computational efficiency.

To effectively ensemble the results from different stages, one na{\"\i}ve approach averages the prediction results obtained by classifiers of different stages during inference. However, the performance is degraded since the classifiers of different stages are not of the same importance. To address this issue, we propose a multistage ensemble loss to cultivate the learning in each stage classifier. Specifically, the multistage ensemble loss forces all stage classifiers to learn from the true label, making each stage classifier's outputs meaningful for our self-ensemble method. We then introduce Dynamic Learner Selection (DLS) to adaptively select the best classifiers based on the training performance of each stage classifier. DLS is a simple non-parametric weighting scalar that has no additional computation. We can apply the DLS to both the ensemble inference and the multistage ensemble loss. The goal of DLS is to maximize the utilization of the best stage classifiers and minimize the stage classifiers with inferior performance. Finally, we illustrate our approach compared with the traditional ensemble for an efficient ensemble in Fig. \ref{fig:problemsetting}. ReSEViT maintains the efficiency of ViT in less data-intensive
domains.

The main research contributions of this study are highlighted as follows:

\begin{enumerate}
  \item To mitigate the Vision Transformer's data-hunger on small image datasets, we provide a new perspective by marrying a multi-stage ViT model with self-ensemble learning via a multi-exit architecture, improving data efficiency. While the traditional methods ensemble multiple models, we present a unified model strategy with residual token and self-ensemble learning for local and global information flows.
  \item To the best of our knowledge, this is the first time self-ensemble learning for a small-data image classification task is proposed. Our ensemble method compresses the training and inference cost by employing the model itself without training additional base models. An efficient DLS method is introduced to dynamically select the best classifiers based on their training performance. Comparison results with other ensemble methods demonstrate the robustness and efficiency of our self-ensemble.
  \item The experimental results show that the proposed ReSEViT is well-motivated for small-scale datasets, including CIFAR, a subset of ImageNet, MedMNIST, and ISIC 2018. Comprehensive studies are conducted to evaluate the effectiveness of ReSEViT across data size, gradient and variance analysis, efficiency, and the importance of residual token with self-ensemble on ReSEViT for small datasets.
\end{enumerate}

\section{Related Works}
\subsection{Transformer in Vision}
The variants of Transformer have not only dominated in natural language processing (NLP) tasks \cite{bert, xlnet}, but also in speech processing \cite{speechformer}, and various vision tasks, such as classification \cite{vit, deit}, object detection \cite{detr}, unsupervised semantic segmentation \cite{unsupervisedsemanticsegmentation}, and human state recognition \cite{husformer}. Specifically, ViT \cite{vit} introduces a pure transformer-based model \cite{transformer} that can attain competitive performance against CNNs \cite{resnet,efficientnet} for image classification on large-scale datasets (\textit{e.g.}, ImageNet-22k and JFT-300M). It partitions an image into fixed-size patches (so-called token embeddings) and adds positional embedding to provide spatial information for each patch embedding. Afterward, the self-attention mechanism is applied to encode the embeddings. However, ViT heavily relies on a large amount of data to learn the global structures of images.

Several works have tried various model architecture designs to better capture the local structures without requiring large amounts of data. For example, DeiT \cite{deit} explores data-efficient training with distillation from other CNN models. T2T-ViT \cite{t2t} overcomes the tokenization problem on ViT by adopting a soft split operation. The soft split operation concatenates multiple tokens within a sliding window into a token, which allows the network to model the local image structures, such as edges and lines. However, the use of multiple tokens concatenation increases computational complexity. As such, PVT \cite{pvt} introduces a progressive spatial reduction technique on attention with multiple stages, which is able to generate feature maps at different scales and reduce computational complexity and memory costs. 

Another way to capture the local structures is to leverage the convolution. For example, CvT \cite{cvt} introduces convolutions on both token embedding and Multihead Self-Attention to model local spatial context for three stages. In this architecture, the positional embedding can be safely removed since the convolution token is able to encode position information for the Transformer. Furthermore, CCT \cite{cct} outperforms CNN-based models by using the single-stage vision transformer architecture (isotropic \cite{convnext}) consisting of a convolution token embedding layer, a transformer layer, and a sequence pooling layer. The sequence pooling layer is used to pool the tokens into a representative token with an attention mechanism in the last layer. Compared with CCT, CvT is a hierarchical three-stage model with local inductive bias applicable to our self-ensemble learning. Henceforth, we extend CvT with residual token embeddings and our self-ensemble methods to aggregate the learned multi-level features obtained from the early into the deepest classifier.

\begin{table*}[!ht]
\renewcommand{\arraystretch}{1.3}
\caption{Comparison Types of Vision Transformer}
\label{table_vit_types}
\centering
\begin{tabular}{c c c c c c c}
\hline
\bfseries Method
& \bfseries Pos. Emb.
& \bfseries Token Emb.
& \bfseries Projections
& \bfseries Multi-Stage
& \bfseries Small-Scale
& \bfseries Efficient Training
\\
\hline
ViT \cite{vit}  & yes & standard & linear & no & no & no
\\
DeiT \cite{deit}  & yes & standard & linear & no & no & distillation
\\
CvT \cite{cvt} & no & convolution & convolution & yes & no & no
\\
CCT \cite{cct} & optional & convolution & linear & no & yes & no
\\
NesT \cite{nest} & yes & convolution & linear & yes & yes & no
\\
TransXNet \cite{transxnet} & yes & convolution & linear & yes & no & no
\\
ReSEViT (Ours)  & no & residual conv. & convolution & yes & yes & self-ensemble
\\
\hline
\end{tabular}
\end{table*}

In summary, we present the key differences between the proposed ReSEViT and other Vision Transformer architectures in Table \ref{table_vit_types}. The table shows the properties of each compared method in terms of positional embeddings, the variety of token embeddings, the variety of projections for attention, transformer stage structure, reliability against a small-scale dataset, and efficient training strategy. As we employ a residual token, a type of convolution, the positional embedding can be removed. Dropping the positional embedding streamlines the computation cost.

\subsection{Ensemble Learning}
The main idea of ensemble learning is to generate multiple weak classifiers and fuse the multiple results, \textit{e.g.}, majority voting, to obtain a better final result~\cite{ensemblesurvey,ensemblefoundation}. The improvements from the fusion of multiple classifiers are due to the following two reasons~\cite{ensemble}. First, if the performance of different classifiers is effective and complementary to each other, the ensemble would fundamentally improve the performance and reduce the risk of a poor prediction. Second, finding the decision hyperplanes for different classes may be too complex for a single classifier; in contrast, ensemble learning effectively finds separation hyperplanes with multiple classifiers.

As such, ensemble methods have been widely adopted in improving the performance of learning tasks, \textit{e.g.}, Bagging~\cite{bagging}, Adaboost~\cite{adaboost}, prediction voting~\cite{ensemblesketchrecognition, ensemblelstm}, and Mixture-of-Experts~\cite{mixturexpert}. Among them, Bagging is one of the earliest ensemble methods, taking bootstrap samples of the training data and learning them independently with several classifiers of the same kind. The individual classifiers are further mixed and voted upon by a simple majority voting scheme for the final decision. In the generation of modern ensemble learning, there is a growing interest in reducing the enormous computation costs associated with fusing multiple neural networks. SnapShot Ensemble (SSE) \cite{snapshot} is an efficient approach that ensembles several sub-models of the same architecture in different local minima by using a cyclical learning schedule rate at no additional training cost. Similar to SSE, Fast Geometric Ensemble (FGE) \cite{fastgeometric} can achieve better performance by finding minima along high-accuracy pathways. Recently, FiLM-Ensemble \cite{filmensemble} introduced a feature-wise linear modulation to construct an ensemble with conditional activation functions under a single monolithic design. However, the monolithic design is slow in practice for mimicking multiple models.

BranchyNet \cite{branchynet} introduced the idea of network-based ensembles and implemented it by introducing multiple exit branches. For each multi-exit branch, joint optimization is applied to prevent overfitting and improve the accuracy of the network. A fixed weighting strategy is further introduced to increase the accuracy by setting a high value for the first branch and a low value for the last branch. As such, they achieve an approximate 1\% accuracy improvement over the uniform weight. However, the fixed weighting strategy may limit its performance. Another related work that uses the idea of self-distillation \cite{selfdistillation} for information boosting has been studied by exploiting the multi-exit architecture. The definition of self-distillation is to transfer knowledge from a teacher part to a student part within the same network. The teacher could be either the deepest stage classifier or the strongest one. Unlike self-distillation, we aim to build a self-ensemble method that integrates different stages instead of forcing every stage to imitate the deepest stage classifier or the strongest one.

\begin{table*}[!ht]
\renewcommand{\arraystretch}{1.3}
\caption{Comparison with early and multi-exit networks}
\label{table_exit_types}
\centering
\resizebox{\textwidth}{!}{
\begin{tabular}{l l l l l}
\hline
\bfseries Method
& \bfseries Training Losses
& \bfseries Inference Phase
& \bfseries Stage Classifier Design
& \bfseries Purpose
\\
\hline
BranchyNet \cite{branchynet} & $\mathcal{L} = \sum^{N}_{n=1} (w_{n} \times \mathcal{L}_{CE}(Z_{n},y))$ (w/ fixed weight) & $Z$ & Extra convolution layers & faster inference
\\
Self-distill \cite{selfdistillation} & $\mathcal{L} = \sum^{N}_{n=1} (\mathcal{L}_{CE}(Z_{n},y) + \mathcal{L}_{distill})$  & $\sum^{N}_{n=1} Z_{n}$ & Extra attention module \& layers & knowledge distillation
\\
Self-ensemble (Ours) & $\mathcal{L} = \sum^{N}_{n=1} (\beta_{n} \times \mathcal{L}_{CE}(Z_{n},y))$ (w/ dynamic weight) & $\sum^{N}_{n=1} \beta^f_{n} \times Z_{n}$ & No extra layer & ensemble
\\
\hline
\end{tabular}
}
\end{table*}

The differences between our self-ensemble method and existing ensemble learning methods are twofold. First, the proposed method ensembles the predictions from different stages of one single model. Second, the proposed method introduces dynamic weighting scalars based on the performance of the training data for better ensemble predictions. A comparison between our approach and existing early and multi-exit networks is shown in Table \ref{table_exit_types}.

\begin{figure*}[pos=ht]
\centering
\includegraphics[scale=0.625]{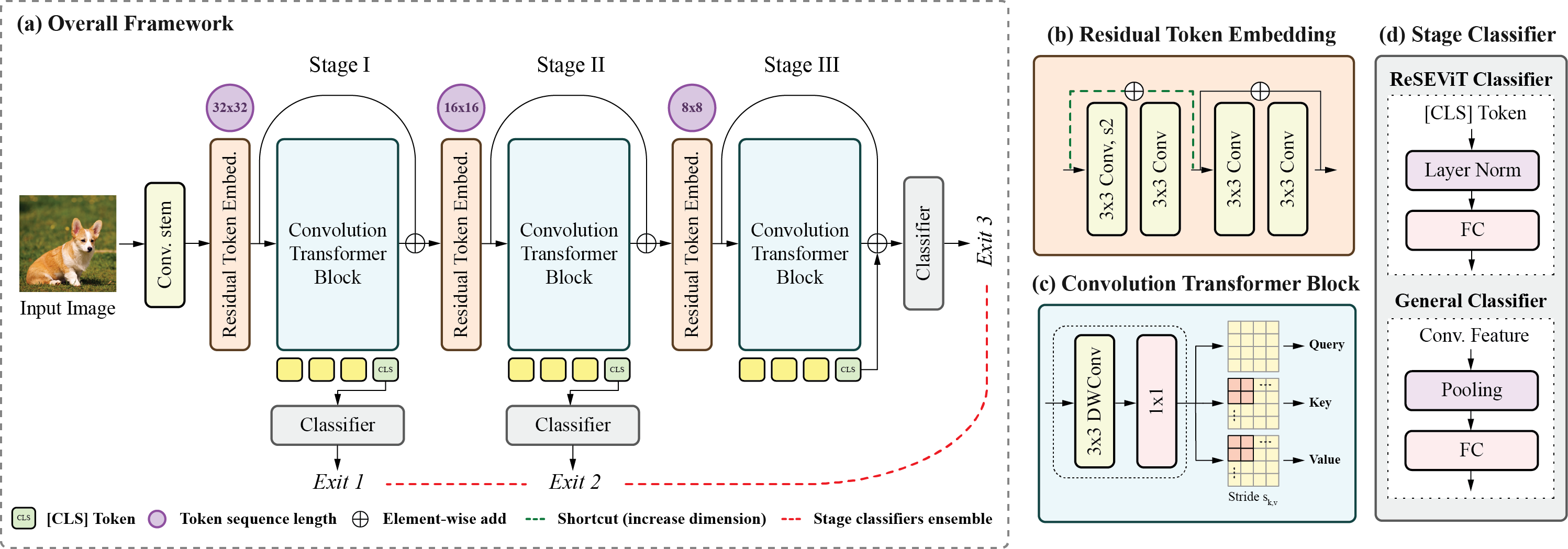}
\caption{(a) Overview of ReSEViT (the multi-exit ViT network architecture with self-ensemble learning for small-scale datasets). ReSEViT is composed of (b) residual token embeddings and (c) convolution transformer blocks. In the residual token embedding, the dashed line denotes a shortcut with an increased dimension (for the second and third-stage), while the solid line is a shortcut with the same dimension. (d) ReSEViT stage classifier consists of a Layer Norm and a FC layer while the general model stage classifier consists of a Pooling and a FC layer.}
\label{fig:model}
\end{figure*}

\begin{figure}[pos=h]
\centering
\includegraphics[scale=0.5]{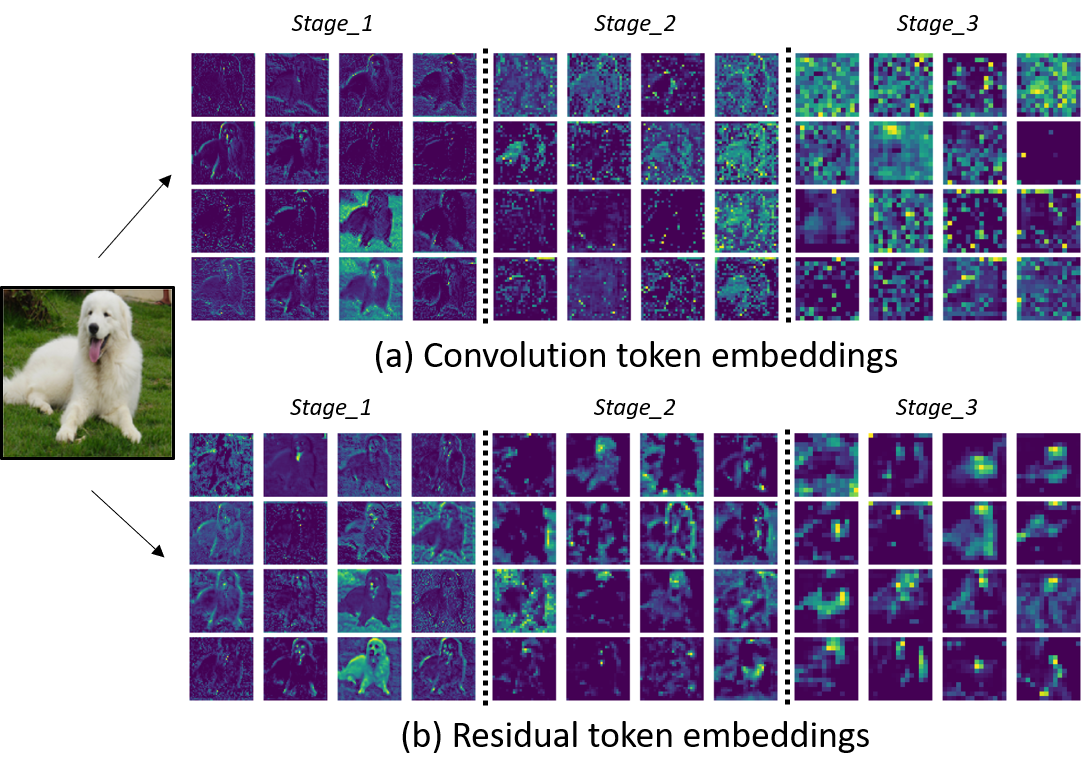}
\caption{Feature visualization of ReSEViT-L model (3 stages) with (a) convolution token; (b) residual token.}
\label{fig:featuremaps}
\end{figure}

\section{Methodology}
\subsection{Model Architecture Overview}
Fig.~\ref{fig:model} shows the architecture of the proposed ReSEViT, which is a three-stage model. Each stage is composed of a residual token embedding and a convolution transformer block \cite{cvt}. In this section, we describe our multi-stage ViT design for small-data image classification tasks. 

\subsubsection{Residual Token Embedding}
We replace the convolution token of CvT \cite{cvt} with the residual token, which is a ResNet block \cite{resnet} for effective local structures where the stronger local representations constitute the locality inductive bias provided to the Transformer. Convolution is an important factor in extracting local structures related to small-scale design in ViT. Generally, convolution layers transform the features by using a deep structure. Given the input image, the first convolutional layer extracts low-level features of the object. When the second or subsequent convolution layer is piled, the features transformed into higher-level features. Based on the convolution output characteristics, the skip connection ensures the network does not forget the lower-level feature. Otherwise, the feature representation becomes too abstract due to deep convolutions. 

In Fig. \ref{fig:featuremaps}, the residual token captures clearer edges, contours, and texture patterns highlighted in stage 1. Moreover, compared with the convolution token, the residual token increasingly focuses on the object and highlights distinct parts, such as the head and body, rather than the background in stage 2. In stage 3, activations in residual token form more coherent patterns corresponding to regions of the object, not only scattered pixels. Based on the learned feature visualization, the convolution layers are employed shown in Eq. \eqref{stacklayers}. 

\begin{equation} 
\label{stacklayers}
	\mathcal{F}(x,\{ W \}) = Conv(Conv(x))
\end{equation}
Afterward, the skip connection is applied to alleviate the optimization difficulty of training a deep network, as shown in Eq. \eqref{skipcon}. 
\begin{equation} 
\label{skipcon}
	y = x + \mathcal{F}(x,\{ W \}), 
\end{equation}
where ${x}$ and ${y}$ are the input and output of the residual block, respectively. ${W}$ represents the parameters of the layers. ${Conv}$ denotes the convolutional layer. $\mathcal{F}$ is the residual function containing $Conv2D$, Batch Normalization, ReLU, $Conv2D$, and Batch Normalization, sequentially.

\subsubsection{Convolution Transformer Block}
 Following \cite{cvt}, the additional local spatial context is modeled on the Multihead Self-Attention (MSA). The token maps are split into Query ($Q$), Key ($K$), and Value ($V$). Next, to provide efficient computation, Depth-Wise Separable convolution (DWS) is adopted for $Q$, $K$, and $V$, as shown in Eq. \eqref{depthwiseQKV}.

\begin{equation} 
\label{depthwiseQKV}
	Q,K,V = DWS(Q), DWS(K), DWS(V).
\end{equation}

 Furthermore, the MSA is computed in Eq. \eqref{transconv}.

\begin{equation} 
\label{transconv}
	MSA = Softmax\bigg(Q K^{T} / \sqrt{d} \bigg) V,
\end{equation}
where $d$ is the scaling factor and $T$ is the token length ($T = H \times W$).

\subsubsection{Stage Classifier Design}
The stage classifier of the proposed ReSEViT is shown in Fig. \ref{fig:model} (d). First, the CLS token is added to each transformer block as a feature representation. The CLS token is then fed to the stage classifier. For general purposes, we also design early-stage classifier architectures for other deep learning models, whose architectures are similar to their deepest-stage classifiers. For example, the ResNet18 stage classifier consists of global max pooling followed by a fully connected (FC) layer.
In general, the entire structure of the stage classifiers remains identical. As such, the additional parameters are from the FC layer of each stage classifier when the self-ensemble method is applied. For efficiency, no convolution layer is involved in all stage classifier designs, including our model and other general models, to explore its effects on the base classifiers.

\subsection{Self-Ensemble Learning}
Similar to the motivation of skip connections, the self-ensemble method ensures that the model prediction result keeps the essential part of the early-stage classifier prediction results if the early-stage classifiers are also supervised during training. The earlier stage classifier prediction results are influenced by the lower-level feature, while the subsequent stage classifier prediction results are influenced by the higher-level feature. Implicitly, the self-ensemble is responsible for global information flow (\textbf{all stages}), while the residual token embedding is responsible for local information flow (\textbf{each individual stage}). In this sense, when the local information flow is stronger, it will be beneficial for the global information flow.

We further propose the self-ensemble method, as shown in the red dashed line of Fig. \ref{fig:model}. Specifically, we ensemble the results from different stages as follows:

\begin{equation}
\label{ensemble}
	k^\star = \argmax^{K}_{k=1} \frac{1}{N} \sum^{N}_{n=1} Z_{n,k},
\end{equation}
where $Z_{n}$ is the $n$-th classifier probability distribution output, $n = 1,..., N$, $k = 1,..., K$, $k^\star$ is the optimal class, $N$ is the number of stage classifiers, and $K$ is the number of classes. In Eq. \eqref{ensemble}, the multiple predictions are averaged, and then $\argmax$ is used to return the maximum value among the class indices for a class decision. 

Theoretically, ViT requires large datasets due to its weak inductive biases, making it prone to overfitting on small datasets. Self-ensemble learning can alleviate this problem by aggregating multiple predictions. Consequently, the prediction variance is reduced, leading to better generalization. Assuming the variance of the self-ensemble can be simplified as:

\begin{equation}
\label{variance}
	Var(\hat{Z}) = \sigma^{2} \bigg(\rho + \frac{1-\rho}{N} \bigg),
\end{equation}

where $\hat{Z}$ is the final ensemble prediction; $\sigma^{2}$ is the variance of an individual prediction; $\rho$ is the average pairwise correlation between ensemble predictions. Eq. \ref{variance} indicates that as $N$ increases and $\rho < 1$, the prediction variance can be reduced, leading to improved generalization, particularly on small datasets, where high variance contributes to overfitting.

\subsubsection{Multistage Ensemble Loss}
A naïve ensemble might degrade the model performance due to the absence of supervision on early-stage classifiers. In other words, the importance of early-stage classifiers is adaptive since they may offer important information in some cases while acting as noise in other cases. To update the model weights for each stage classifier directly from the true label, we propose the multistage ensemble loss. Given training samples $X = \{ x,y \}$, where $x$ is the input image, and $y$ is the true label, the probability distribution output of each stage classifier is supervised by the true label $y$, \textit{i.e.},

\begin{equation} 
\label{stageloss}
	\mathcal{L}_{SE} = \frac{1}{N} \sum^{N}_{n=1} \mathcal{L}_{CE}(Z_{n},y),
\end{equation}
where $\mathcal{L}_{SE}$ represents self-ensemble loss and $\mathcal{L}_{CE}$ is the label smoothing cross-entropy loss.

\subsubsection{Dynamic Learner Selection (DLS)}
With the multistage ensemble loss, the early-stage classifiers perform better than a setting that uses a single loss. However, we argue that the outputs of different stages should not have the same importance. Therefore, we further propose Dynamic Learner Selection (DLS) to adaptively adjust the stage-classifier weights. The goal of the DLS is to maximize the use of best stage classifiers while minimizing the stage classifiers with an inferior performance by adding priority scores $\beta$, which is based on the training accuracy over epochs of each stage classifier. It is simple and effective with no additional parameters. Concretely, let $\beta_n$ and $acc_{n}$ denote the weight of the $n$-th stage classifier and the training accuracy of $Z_{n}$, respectively. $\beta_n$ can be calculated as follows:

\begin{equation} 
\label{beta}
	\beta_{n} = \frac{acc_{n}}{\sum^{N}_{n=1} acc_{n}+\epsilon}.
\end{equation} 	

To obtain the priority score, we normalize the $acc_{n}$ by summarizing the accuracy of all stage classifiers and add a very small value $\epsilon$ to Eq.~\eqref{beta} to prevent the priority score from being infinite in the early epoch. The priority score is in the range [0,1]. The new loss function weighted self-ensemble loss $\mathcal{L}_{WSE}$ and prediction with DLS can be calculated by:

\begin{equation}
\label{multistageloss_dynamic}
	\mathcal{L}_{WSE} =  \sum^{N}_{n=1} \beta_{n} \times \mathcal{L}_{CE}(Z_{n},y).
\end{equation}

\begin{equation} 
\label{ensemble_dynamic}
	k^\star = \argmax^{K}_{k=1} \big(  \sum^{N}_{n=1} \beta^f_{n} \times Z_{n,k} \big).
\end{equation}

Note that $\beta$ is sequentially updated for every training epoch. In the training process, we calculated the $\beta$ values and used them to modify the multistage ensemble loss. After training, we have the final result of $\beta$, denoted by $\beta^f$, and then apply the $\beta^f$ to the ensemble classifier for the inference.

\begin{table*}[ht]
\renewcommand{\arraystretch}{1.3}
\caption{Detailed Settings of ReSEViT Variants.}
\label{table_architecture}
\centering
\resizebox{\textwidth}{!}{
\begin{tabular}{c|l|c|c|c|c}
    \hline
    \multicolumn{4}{c|}{\bfseries Small Resolution (32$\times$32)} & \multicolumn{2}{c}{\bfseries Large Resolution (224$\times$224)}
    \\
    \hline
    \bfseries Output & \bfseries Layer Name & \bfseries ReSEViT-T & \bfseries ReSEViT-S & \bfseries Output & \bfseries ReSEViT-L \\
    \hline 
    32\texttimes32 & 
    Conv. Stem & 
    3\texttimes3, 64, $s$=1 &
    3\texttimes3, 64, $s$=1 &
    56\texttimes56 &
    7\texttimes7, 64, $s$=4
    \\
    \hline
    \multirow{2}{*}{32\texttimes32} & 
    \multirow{2}{*}{Res. Token 1} & 
    \multirow{2}{*}{$\Big[$
    \makecell{
        3\texttimes3, 64, $s$=1 \\
        3\texttimes3, 64, $s$=1
    }
    $\Big]\times2$} 
    & 
    \multirow{2}{*}{$\Big[$
    \makecell{
        3\texttimes3, 64, $s$=1 \\
        3\texttimes3, 64, $s$=1
    }
    $\Big]\times2$} 
    & 
    \multirow{2}{*}{56\texttimes56}
    &
    \multirow{2}{*}{$\Big[$
    \makecell{
        3\texttimes3, 64, $s$=1 \\
        3\texttimes3, 64, $s$=1
    }
    $\Big]\times2$} 
    
    \\
    & & & & &
    \\
    
    \hline
    \multirow{2}{*}{32\texttimes32} & 
    \multirow{2}{*}{Transformer Block 1} & 
    \multirow{2}{*}{$\Big[$
    \makecell{
        3\texttimes3, 64, $s_{k,v}$=2 \\
        $T_{1}$=32, $h_{1}$=4
    }
    $\Big]\times1$} 
    & 
    \multirow{2}{*}{$\Big[$
    \makecell{
        3\texttimes3, 64, $s_{k,v}$=1 \\
        $T_{1}$=32, $h_{1}$=8
    }
    $\Big]\times2$} 
    & 
    \multirow{2}{*}{56\texttimes56}
    &
    \multirow{2}{*}{$\Big[$
    \makecell{
        3\texttimes3, 64, $s_{k,v}$=2 \\
        $T_{1}$=56, $h_{1}$=1
    }
    $\Big]\times2$} 
    \\
    & & & & &
    \\
    
    \hline
    \multirow{2}{*}{16\texttimes16} & 
    \multirow{2}{*}{Res. Token 2} & 
    \multirow{2}{*}{$\Big[$
    \makecell{
        3\texttimes3, 128, $s$=2 \\
        3\texttimes3, 128, $s$=1
    }
    $\Big]\times2$}
    & 
    \multirow{2}{*}{$\Big[$
    \makecell{
        3\texttimes3, 128, $s$=2 \\
        3\texttimes3, 128, $s$=1
    }
    $\Big]\times2$}
    & 
    \multirow{2}{*}{28\texttimes28}
    &
    \multirow{2}{*}{$\Big[$
    \makecell{
        3\texttimes3, 192, $s$=2 \\
        3\texttimes3, 192, $s$=1
    }
    $\Big]\times2$} 
    \\
    & & & & &
    \\
    
    \hline
    \multirow{2}{*}{16\texttimes16} & 
    \multirow{2}{*}{Transformer Block 2} & 
    \multirow{2}{*}{$\Big[$
    \makecell{
        3\texttimes3, 128, $s_{k,v}$=2 \\
        $T_{2}$=16, $h_{2}$=4
    }
    $\Big]\times1$} 
    &
    \multirow{2}{*}{$\Big[$
    \makecell{
        3\texttimes3, 128, $s_{k,v}$=1 \\
        $T_{2}$=16, $h_{2}$=8
    }
    $\Big]\times2$}
    & 
    \multirow{2}{*}{28\texttimes28}
    &
    \multirow{2}{*}{$\Big[$
    \makecell{
        3\texttimes3, 192, $s_{k,v}$=2 \\
        $T_{2}$=28, $h_{2}$=3
    }
    $\Big]\times2$}
    \\
    & & & & &
    \\
    
    \hline
    \multirow{2}{*}{8\texttimes8} & 
    \multirow{2}{*}{Res. Token 3} & 
    \multirow{2}{*}{$\Big[$
    \makecell{
        3\texttimes3, 256, $s$=2 \\
        3\texttimes3, 256, $s$=1
    }
    $\Big]\times2$} 
    & 
    \multirow{2}{*}{$\Big[$
    \makecell{
        3\texttimes3, 256, $s$=2 \\
        3\texttimes3, 256, $s$=1
    }
    $\Big]\times2$} 
    & 
    \multirow{2}{*}{14\texttimes14}
    &
    \multirow{2}{*}{$\Big[$
    \makecell{
        3\texttimes3, 384, $s$=2 \\
        3\texttimes3, 384, $s$=1
    }
    $\Big]\times2$} 
    \\
    & & & & &
    \\
    
    \hline
    \multirow{2}{*}{8\texttimes8} & 
    \multirow{2}{*}{Transformer Block 3} &
    \multirow{2}{*}{$\Big[$
    \makecell{
        3\texttimes3, 256, $s_{k,v}$=2 \\
        $T_{3}$=8, $h_{3}$=4
    }
    $\Big]\times1$} 
    &
    \multirow{2}{*}{$\Big[$
    \makecell{
        3\texttimes3, 256, $s_{k,v}$=1 \\
        $T_{3}$=8, $h_{3}$=8
    }
    $\Big]\times1$} 
    & 
    \multirow{2}{*}{14\texttimes14}
    &
    \multirow{2}{*}{$\Big[$
    \makecell{
        3\texttimes3, 384, $s_{k,v}$=2 \\
        $T_{3}$=14, $h_{3}$=6
    }
    $\Big]\times3$} 
    \\
    & & & & &
    \\
    \hline
    
    1\texttimes1 & 
    Linear &
    $100$ &
    $100$ &
    1\texttimes1
    &
    $100$
    \\
    
    \hline
    \multicolumn{2}{c|}{\textbf{(Params, FLOPs)}} &  (3.88M, 0.56G) & (4.13M, 0.68G) & & (12.43M, 4.08G)
    \\
    \hline
\end{tabular}
}
\end{table*} 

\section{Experiments}
\subsection{Experimental Setup}

\begin{table}[H]
\renewcommand{\arraystretch}{1.3}
\caption{Statistics of Small-Scale Datasets.}
\label{table_data}
\centering
\begin{tabular}{lccccc}
\toprule
\bfseries Dataset
& \bfseries Train
& \bfseries Test
& \bfseries Dimension
& \bfseries{\# Classes}
\\
\midrule
\multicolumn{5}{c}{\bfseries{General Datasets}}
\\
CIFAR10 \cite{cifar} & 50,000 & 10,000 & 32$\times$32 & 10
\\
CIFAR100 \cite{cifar} & 50,000 & 10,000 & 32$\times$32 & 100
\\
Tiny-ImageNet \cite{tinyimagenet} & 100,000 & 10,000 & 64$\times$64 & 200
\\
ImageNet100 \cite{cmc} & 130,000 & 5,000 & 224$\times$224 & 100
\\
\midrule
\multicolumn{5}{c}{\bfseries{Medical Datasets}}
\\
MedMNIST \cite{medmnistv1} & $\sim$546–97K & $\sim$78–11K & 28$\times$28 & 2-14
\\
ISIC 2018 \cite{isic} & 8,012 & 2,003 & 224$\times$224 & 7
\\
\hline
\end{tabular}
\end{table}

\noindent\textbf{Datasets.} As shown in Table \ref{table_data}, we conduct the following experiments on small-scale general datasets benchmark, \textit{i.e.}, CIFAR10, CIFAR100, and Tiny-ImageNet. Tiny-ImageNet is a subset of the original ImageNet. As the small-scale datasets cover both small and large image resolution, we also conduct our experiments on ImageNet100, a subset of the ImageNet1K dataset \cite{imagenet}. MedMNIST, a small-scale medical imaging dataset comprising 10 subsets with a 28$\times$28 image resolution, is also used for evaluation. Moreover, ISIC 2018 is a skin lesion dataset containing 10,015 labeled images (450$\times$600 resolution). The dataset is split into 80\% training and 20\% testing, and all images are resized to 224$\times$224.

\noindent\textbf{Model Variants.} We develop three variants of ReSEViT on small-scale datasets with different sizes, such as ReSEViT-T (tiny), ReSEViT-S (small), and ReSEViT-L (large), as shown in Table \ref{table_architecture}. Both ReSEViT-T and ReSEViT-S are evaluated on small-resolution image datasets. On the other hand, ReSEViT-L is designed for a large-resolution image dataset. Three stages are employed. $h_{n}$ denotes the head number of each stage, $s$ is the stride of convolution, $s_{k,v}$ denotes the stride of key and value, and $T_{n}$ is the token sequence length.

\noindent\textbf{Implementation Details.} During training, AdamW \cite{adamW} is used as the optimizer with a weight decay of 3e-2. We train our models for 200 epochs with a batch size of 64, an initial learning rate of 5e-4, a cosine learning rate scheduler \cite{cosinelr}, and label smoothing with a probability of 0.1. While training, our methods are warmed up for 10 epochs. For data augmentation, we use horizontal flipping, random cropping, and AutoAugment \cite{autoaugment}. For Tiny-ImageNet, the stride of the convolutional stem in all models is doubly increased (from 1 to 2) to accommodate the 32$\times$32 receptive field, hence, the computational complexity remains equivalent. For ImageNet100 and ISIC 2018, we train from scratch for 100 epochs with a batch size of 64. All experimental results are averaged over three runs. For comparison, prior method results are reported or trained with our training setting when the official implementation is publicly available.

\begin{table*}
\renewcommand{\arraystretch}{1.3}
\caption{Test accuracy comparison on small-scale image datasets with small resolution (32$\times$32). $^\dagger$ represents a modified stem for small resolution datasets, while $^\star$ represents the original architecture without modification. We report accuracy, parameter size, FLOPs, GPU training memory usage, and inference throughput (images/s).}
\label{table_sota}
\centering
\resizebox{\columnwidth}{!}{
\begin{tabular}{lcccc|cccc}
\hline
\bfseries Method
& \bfseries CIFAR10
& \bfseries CIFAR100
& \bfseries Tiny-ImageNet
& \bfseries Avg. Acc $\uparrow$
& \bfseries Params. $\downarrow$
& \bfseries FLOPs $\downarrow$
& \bfseries GPU $\downarrow$
& \bfseries Throughput $\uparrow$
\\
\hline
\multicolumn{9}{c}{Convolution Networks}
\\
\hline
ResNet18$^\dagger$ \cite{resnet} & 94.37\% & 75.41\% & 52.47\% & 74.08\% & 11.22M & 0.56G & 2045M & 10351 
\\
ResNet34$^\dagger$ \cite{resnet} & 95.26\% & 77.08\% & 54.87\% & 75.74\% & 21.33M & 1.16G & 2081M & 7360
\\
DenseNet 100-12 \cite{densenet} & 95.90\% & 79.80\% & 54.81\% & 76.84\% & 7.00M & 1.85G & 5347M & 1251
\\
WRN-40-4 \cite{wideresnet} & 95.47\% & 78.90\% & 56.86\% & 77.08\% & 8.98M & 1.42G & 2035M & 5812
\\
ConvNeXt-T$^\dagger$ \cite{convnext} & 94.55\% & 74.02\% & 51.15\% & 73.24\% & 27.94M & 1.45G & 4446M & 3455
\\
Conv2Former-N$^\dagger$ \cite{conv2former} & 94.37\% & 77.19\% & 57.24\% & 76.27\% & 13.83M & 0.73G & 5843M & 3430
\\
\hline
\multicolumn{9}{c}{Vision Transformers}
\\
\hline
ViT-B \cite{vit} & 76.42\% & 46.61\% &  25.21\% & 49.41\% & 85.63M & 0.43G & 3339M & 6721
\\
ViT-B (4$\times$4 patch size) \cite{vit} & 82.68\% & 65.72\% & 44.52\% & 64.31\% & 85.19M & 5.52G & 6217M & 1820
\\
CvT-13$^\star$ \cite{cvt} & 88.45\% & 58.48\% & 46.28\% & 64.40\% & 19.66M & 0.10G & 2147M & 2781
\\
CvT-13$^\dagger$ \cite{cvt} & 94.51\% & 73.56\% & 53.28\% & 73.78\% & 19.65M & 1.32G & 4247M & 2560
\\
CCT-7/3$\times$1 \cite{nest} & 94.72\% & 76.67\% & 49.42\% & 73.60\% & 3.76M & 1.19G & 4487M & 3140 
\\
Swin-T$^\dagger$ \cite{nest} & 94.46\% & 78.07\% & 52.12\% & 74.88\% & 27.5M & 1.95G & 5560M & 1850
\\
PoolFormer-S36$^\dagger$ \cite{poolformer} & 94.81\% & 76.44\% & 55.81\% & 75.69\% & 30.39M & 1.62G & 5721M & 1930
\\
NesT-T \cite{nest} & 96.04\% & 78.69\% & 56.17\% & 76.97\% & 6.28M & 2.80G & 6870M & 1655
\\
TransXNet-T$^\dagger$ \cite{transxnet} & 94.77\% & 75.33\% & 54.33\% & 74.81\% & 12.17M & 0.58G & 6962M & 2155
\\
\grayline
ReSEViT-T (w/o $\mathcal{L}_{WSE}$) & 95.10\% & 77.65\% & 56.05\% & 76.27\% & 3.86M & 0.56G & 2380M & 4841
\\
\rowcolor{Gray}
\bfseries{ReSEViT-T (Ours)} & 95.60\% & 79.49\% & 60.07\% & 78.39\% & 3.88M & 0.56G & 2397M & 4832
\\
ReSEViT-S (w/o $\mathcal{L}_{WSE}$) & 95.84\% & 78.65\% & 57.42\% & 77.30\% & 4.11M & 0.68G & 6030M & 1431
\\
\rowcolor{Gray}
\bfseries{ReSEViT-S (Ours)} & 96.17\% & 81.05\% & 61.75\% & \bfseries{79.65\%} & 4.13M & 0.68G & 6047M & 1415
\\
\hline
\end{tabular}
}
\end{table*}

\subsection{Small-Data Image Classification}
\subsubsection{Training on Small Resolution Data}
We evaluate the performance of our ReSEViT on small-resolution image datasets. Table \ref{table_sota} compares our approach with CNN-based and ViT models in various aspects, \textit{i.e.}, accuracy and efficiency. 

\noindent\textbf{ViT:} Based on the result, we can see that ViT-B underperforms with its pure transformer architecture. Since the input size is 32$\times$32, we report another different patch size from 16$\times$16 to 4$\times$4, resulting in a quadratic increase in the FLOPs (12.8$\times$) and low inference throughput (3.7$\times$). This is because the number of tokens directly affects the input sequence length from 4 to 64 to the transformer layers. The ViT result shows that smaller patch sizes for the transformer still cannot really compete with the hybrid ViT in a light computational budget with a limited amount of data, indicating a data-hunger.

\noindent\textbf{Hybrid ViT:} Compared with CvT-13$^\dagger$ as the hybrid ViT base form, our ReSEViT-T achieves 4.61\% better performance while improving its efficiency because the residual token and self-ensemble method suggest local and global information flows. NesT-T performs better than the CvT-13$^\dagger$, CCT-7/3$\times$1, and TransXNet-T$^\dagger$ in accuracy performance with the nested architecture and local translations, but the throughput speed is much slower. Our ReSEViT-T achieves better performance than NesT-T by 1.42\%, while the model size is more compact and the throughput is faster. This shows the effectiveness of our approach on small-scale datasets.

\begin{table}[H]
\renewcommand{\arraystretch}{1.3}
\caption{Test accuracy comparison on ImageNet100 (224$\times$224). We report accuracy, parameter size, FLOPs, and inference throughput (images/s). $^\ddagger$ applies pre-trained RegNetY-16GF from ImageNet-1K hard distillation \cite{efficientformer} with top-1 88.46$\%$ accuracy on ImageNet100.}
\label{table_largescale}
\centering
\begin{tabular}{lcccc}
\hline
\bfseries Method
& \bfseries Top-1 $\uparrow$
& \bfseries Params. $\downarrow$
& \bfseries FLOPs $\downarrow$
& \bfseries Throughput $\uparrow$
\\
\hline
ViT-B (ICLR'21) \cite{vit} & 71.38\% & 85.85M & 16.85G & 605
\\
T2T-ViT-14 (ICCV'21) \cite{t2t} & 78.48\% & 21.12M & 4.33G & 1170
\\
Swin-T (ICCV'21)  \cite{swin}& 80.66\% & 27.60M & 4.35G & 930
\\
CvT-13 (ICCV'21)  \cite{cvt}& 81.57\% & 19.69M & 4.05G & 1034
\\
EfficientFormer-L3 (NeurIPS'22) \cite{efficientformer} & 79.21\% & 30.48M & 3.95G & 916
\\
PoolFormer-S36 (CVPR'22) \cite{poolformer}& 81.25\% & 30.40M & 4.99G & 680
\\
EfficientFormer-L3$^\ddagger$ (NeurIPS'22) \cite{efficientformer} & 82.49\% & 30.48M & 3.95G & 916
\\
ConvNeXt-T (CVPR'22) \cite{convnext} & 83.35\% & 27.97M & 4.46G & 1208
\\
CrossFormer-S (ICLR'22) \cite{crossformer} & 83.55\% & 29.97M & 4.77G & 851
\\
NesT-T (AAAI'22) \cite{nest} & 84.36\% & 16.71M & 5.22G & 784
\\
FasterNet-M (CVPR'23) \cite{fasternet} & 82.73\% & 52.37M & 8.75G & 649
\\
ViTAEv2-S (IJCV'23) \cite{vitaev2} & 84.03\% & 19.36M & 5.50G & 385
\\
CrossFormer++-S (TPAMI'23) \cite{crossformer++} & 84.87\% & 22.86M & 4.05G & 716
\\
Conv2Former-N (TPAMI'24) \cite{conv2former}  & 85.05\% & 13.84M & 2.37G & 1074
\\
TransXNet-T (TNNLS'25) \cite{transxnet} & 82.97\% & 12.43M & 1.78G & 792
\\
CAT-T (AAAI'26) \cite{cat} & 84.24\% & 26.22M & 3.40G & 1010
\\
\rowcolor{Gray}
\bfseries{ReSEViT-L (Ours)} & \textbf{85.23\%} & 12.43M & 4.08G & 890
\\
\hline
\end{tabular}
\end{table}

\subsubsection{Training on Large Resolution Data}
In Table \ref{table_largescale}, we evaluate our performance result of ReSEViT on a large resolution image dataset, ImageNet100, which is 10$\%$ of the ImageNet1K dataset, to analyze the ViT performance under limited data. The result shows that ViT-B remains inferior in accuracy and efficiency to those of hybrid ViTs. Compared with our base form model, CvT-13, ReSEViT-L achieves 3.66\% better performance with less parameter size. Moreover, ReSEViT-L also behaves better than NesT-T and TransXNet-T in accuracy with a comparable computational budget.

\begin{table}[!ht]
\renewcommand{\arraystretch}{1.3}
\caption{Comparative performance of ViT variants across MedMNIST sub-datasets. AUC and Top-1 accuracy are reported.}
\label{table_medical}
\centering
\begin{tabular}{ccccca}
\hline
\bfseries Datasets
& \bfseries Metrics
& \bfseries CvT-13$^\dagger$
& \bfseries CCT-7/3$\times$1
& \bfseries NesT-T
& \bfseries ReSEViT-T
\\
\hline
\multirow{2}{*}{PathMNIST} & AUC & 0.9951 & 0.9953 & 0.9956 & \textbf{0.9958}
\\
 & Top-1 & 0.9560 & 0.9554 & 0.9620 & \textbf{0.9692}
\\
\hline
\multirow{2}{*}{ChestMNIST} & AUC & 0.5742 & 0.4764 & 0.5679 & \textbf{0.5849}
\\
 & Top-1 & \textbf{0.9496} & 0.9493 & 0.9495 & \textbf{0.9496}
\\
\hline
\multirow{2}{*}{DermaMNIST} & AUC & 0.7356 & 0.7619 & 0.6591 & \textbf{0.7645}
\\
 & Top-1 & 0.7292 & 0.7360 & 0.6686 & \textbf{0.7399}
\\
\hline
\multirow{2}{*}{OCTMNIST} & AUC & 0.9839 & 0.9840 & \textbf{0.9855} & 0.9850
\\
 & Top-1 & \textbf{0.9549} & 0.9495 & 0.9545 & 0.9534
\\
\hline
\multirow{2}{*}{PneumoniaMNIST} & AUC & 0.9981 & 0.9977 & 0.9967 & \textbf{0.9993}
\\
 & Top-1 & 0.9844 & 0.9664 & 0.9740 & \textbf{0.9878}
\\
\hline
\multirow{2}{*}{RetinaMNIST} & AUC & 0.6800 & 0.7402 & 0.6916 & \textbf{0.7723}
\\
 & Top-1 & 0.4453 & \textbf{0.4732} & 0.4453 & 0.4609
\\
\hline
\multirow{2}{*}{BreastMNIST} & AUC & 0.8586 & 0.9707 & 0.8886 & \textbf{0.9714}
\\
 & Top-1 & 0.8270 & \textbf{0.9375} & 0.8862 & 0.9252
\\
\hline
\multirow{2}{*}{Organ\_A\_MNIST} & AUC & \textbf{0.9706} & \textbf{0.9706} & \textbf{0.9706} & \textbf{0.9706}
\\
 & Top-1 & 0.9943 & 0.9933 & \textbf{0.9957} & 0.9944
\\
\hline
\multirow{2}{*}{Organ\_C\_MNIST} & AUC & 0.9804 & 0.9804 & \textbf{0.9808} & \textbf{0.9808}
\\
 & Top-1 & 0.9757 & 0.9755 & 0.9836 & \textbf{0.9864}
\\
\hline
\multirow{2}{*}{Organ\_S\_MNIST} & AUC & 0.9713 & 0.9709 & 0.9714 & \textbf{0.9718}
\\
 & Top-1 & 0.9126 & 0.8994 & \textbf{0.9137} & 0.9117
\\
\hline
\multirow{2}{*}{Average} & AUC & 0.8748 & 0.8848 & 0.8708 & \textbf{0.8996}
\\
 & Top-1 & 0.8729 & 0.8836 & 0.8733 & \textbf{0.8879}
\\
\hline
\end{tabular}
\end{table}

\begin{table}
\renewcommand{\arraystretch}{1.3}
\caption{Comparative performance of ViT variants on ISIC 2018 (224$\times$224). AUC and Top-1 Accuracy are reported.}
\label{table_medicalisic}
\centering
\begin{tabular}{lcccc|cca}
\hline
\bfseries \multirow{2}{*}{Metrics}
& \bfseries \multirow{2}{*}{ViT-B}
& \bfseries \multirow{2}{*}{CvT-13}
& \bfseries \multirow{2}{*}{CCT-14/7$\times$2}
& \bfseries \multirow{2}{*}{NesT-T}
& \multicolumn{3}{c}{\bfseries ReSEViT-L}
\\
 & & & & & SL & ML & \bfseries{ML+DLS} 
\\
\hline
AUC & 0.8410 & 0.8766 & 0.7728 & 0.8798 & 0.8972 & 0.9113 & \textbf{0.9176}
\\
Top-1 & 0.5837 & 0.6407 & 0.6733 & 0.6483 & 0.6707 & 0.6895 & \textbf{0.7066}
\\
\hline
\end{tabular}
\end{table}

\subsubsection{Training on Real-World Medical Imaging Data}
Table \ref{table_medical} shows the performance comparison of our approach and other ViT variants on small-scale medical image datasets. Area Under the Curve (AUC) and Top-1 accuracy are chosen as metrics due to the common class imbalance issue present in the MedMNIST sub-datasets \cite{smalldatasetaugmentation}. In this experiment, the model with the highest AUC on the validation set is selected for the test. Based on the results, we observe that ReSEViT-T outperforms the other ViT variants in the small-scale medical domain. Besides the MedMNIST benchmark, we also conducted a comparative evaluation of ViT variants on the ISIC 2018 image dataset in Table \ref{table_medicalisic}, where the proposed approach attained the highest AUC and accuracy among the evaluated ViT models. The experimental results also confirm that the self-ensemble components, particularly DLS, effectively enhance the overall performance.

\begin{figure}[pos=!h]
\centering
\includegraphics[scale=0.5]{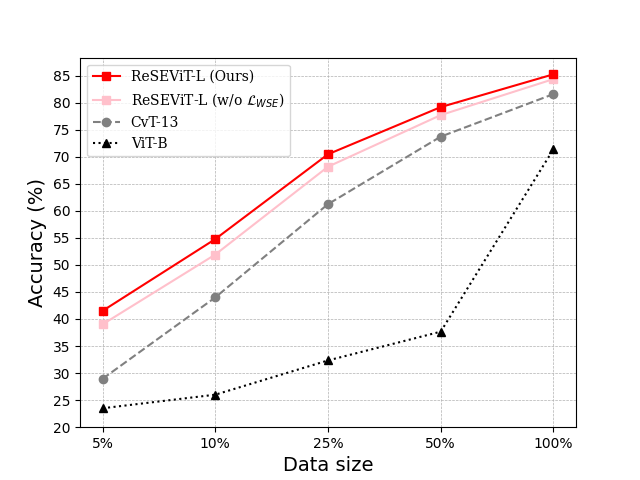}
\caption{Comparison of ViTs against data size on ImageNet100.}
\label{fig:datasize}
\end{figure}

\subsection{Impact of Data Size on Vision Transformer} 
Vision Transformer faces data-hungry problems when the training data is limited, also known as small-scale datasets. In this part, we explore how training data size impacts ViT's performance. An experiment is conducted on the ImageNet100 dataset as a benchmark to observe the correlation between the amount of training data and the accuracy achieved.

Based on the experimental result in Fig. \ref{fig:datasize}, hybrid ViTs are able to perform better than ViT-B under limited data. Concretely, the trend of the results tends to be convex upwards, while the ViT-B's results are concave downwards. Our ReSEViT-L achieves 79.17\% with only 50\% of the data, while ViT-B achieves 37.66\%. It is also observed that ReSEViT-L outperforms the CvT-13 as our base form architecture by a very large margin. This shows that our approach is less data-hungry than ViT-B and CvT-13 by incorporating residual tokens and self-ensemble learning.

\begin{table}[H]
\renewcommand{\arraystretch}{1.3}
\caption{Comparison with Other Ensemble Methods in ReSEViT-S on CIFAR100. $M$ is the number of ensemble models. $^{	\blacklozenge}$ refers to a parallel ensemble type.}
\label{table_ensemblecompare}
\centering
\begin{tabular}{lcccccc}
\hline
\bfseries Ensemble Method
& \bfseries $M$ 
& \bfseries Acc $\uparrow$
& \bfseries Params. $\downarrow$
& \bfseries Train $\downarrow$
& \bfseries Test $\downarrow$
\\
\hline
Baseline & 1 & 78.65\% & 4.11M & 1.00$\times$ & 1.00$\times$ 
\\
Model Avg. & 2 & 79.89\% & 13.74M$^{\blacklozenge}$ & 1.05$\times$ & 1.02$\times$ 
\\
Bagging \cite{bagging} & 2 & 78.76\% & 8.22M$^{\blacklozenge}$ & 2.41$\times$ & 1.05$\times$ 
\\
SSE \cite{snapshot} & 4 & 79.41\% & 4.11M & 1.00$\times$ & 1.17$\times$ 
\\
FGE \cite{fastgeometric} & 12 & 79.51\% & 4.11M & 1.00$\times$ & 2.86$\times$ 
\\
FiLM \cite{filmensemble} & 2 & 80.07\% & 4.12M & 2.18$\times$ & 1.6$\times$ 
\\
\rowcolor{Gray}
\bfseries{Self-ensemble (Ours)} & 1 & \textbf{81.05\%} & 4.13M & 1.01$\times$ & 1.01$\times$ 
\\
\hline
\end{tabular}
\end{table}

\subsection{Comparison with Other Ensemble Methods} 
Table \ref{table_ensemblecompare} shows the comparison with other ensemble methods on the CIFAR100 dataset. Specifically, the accuracy, parameter size budget, and proportion speed of training and inference are reported to validate the efficiency and effectiveness of our approach. The inference time is taken from the GPU time to run a forward pass for a batch size of 64. We use the same training settings for all ensemble methods, including batch size, learning rate, loss functions, and number of epochs for a fair comparison. Various ensemble method types are presented, such as parallel (Model Averaging and Bagging), sequential (SSE and FGE), implicit ensemble (FiLM-Ensemble), and our proposed self-ensemble learning. For comparison, we use the ReSEViT-S without the self-ensemble as the baseline model. Further, we describe each ensemble method and make the following observations:
\begin{itemize}
\item \textbf{Model Averaging.} Two models are separately trained, \textit{i.e.}, Plain-ReSEViT-S (ReSEViT-S w/o residual token and self-ensemble) and ResNet18. After training, the predictions of the deepest classifiers are averaged. The model averaging method improves by 2.61\% compared with Plain-ReSEViT-S (77.28\%) and improves by 4.48\% compared with ResNet18 (75.41\%). Moreover, the result shows that our ReSEViT-S achieves 1.16\% better accuracy than the model averaging method in just one model with fewer parameters and efficient speed. Based on the result, merging two models into a unified model is more effective than using two individual distinct models.
\item \textbf{Bagging.} The bagging ensemble of two baseline models is trained. This approach only improves the baseline model by 0.11\% and has a high computational cost since two identical models are used in parallel.
\item \textbf{SSE.} The baseline model is used with the SSE technique of four-cycle models (50 epochs per cycle). The method improves the baseline model by 0.76\% because the model states sequentially taken during the training are highly correlated. In contrast, our self-ensemble method has a high diversity representation in each stage classifier, leading to an increase.
\item \textbf{FGE.} A pre-training routine for 100 epochs and a cycle length of 4 are used. We set the initial learning rate of 5e-4 and the smallest learning rate of 1e-6. The total number of models in the final process is 12. The result shows that FGE performs slightly better than SSE, while the speed in the inference is much slower since a lot of ensemble models are used to compensate for the accuracy performance. 
\item \textbf{FiLM-Ensemble.} We train the baseline model with two ensemble members. The method improves the baseline model by 1.42\% and is better than the other ensemble methods. However, the main drawbacks of FiLM-Ensemble are the training and inference speed, based on the results, since a monolithic design is used. We conjecture that repeating the batch in the training process to train on more data and reshaping back the repeated batch in the inference process slows the speed.
\end{itemize}

\begin{figure*}[pos=t]
 \centering
 \includegraphics[scale=0.495]{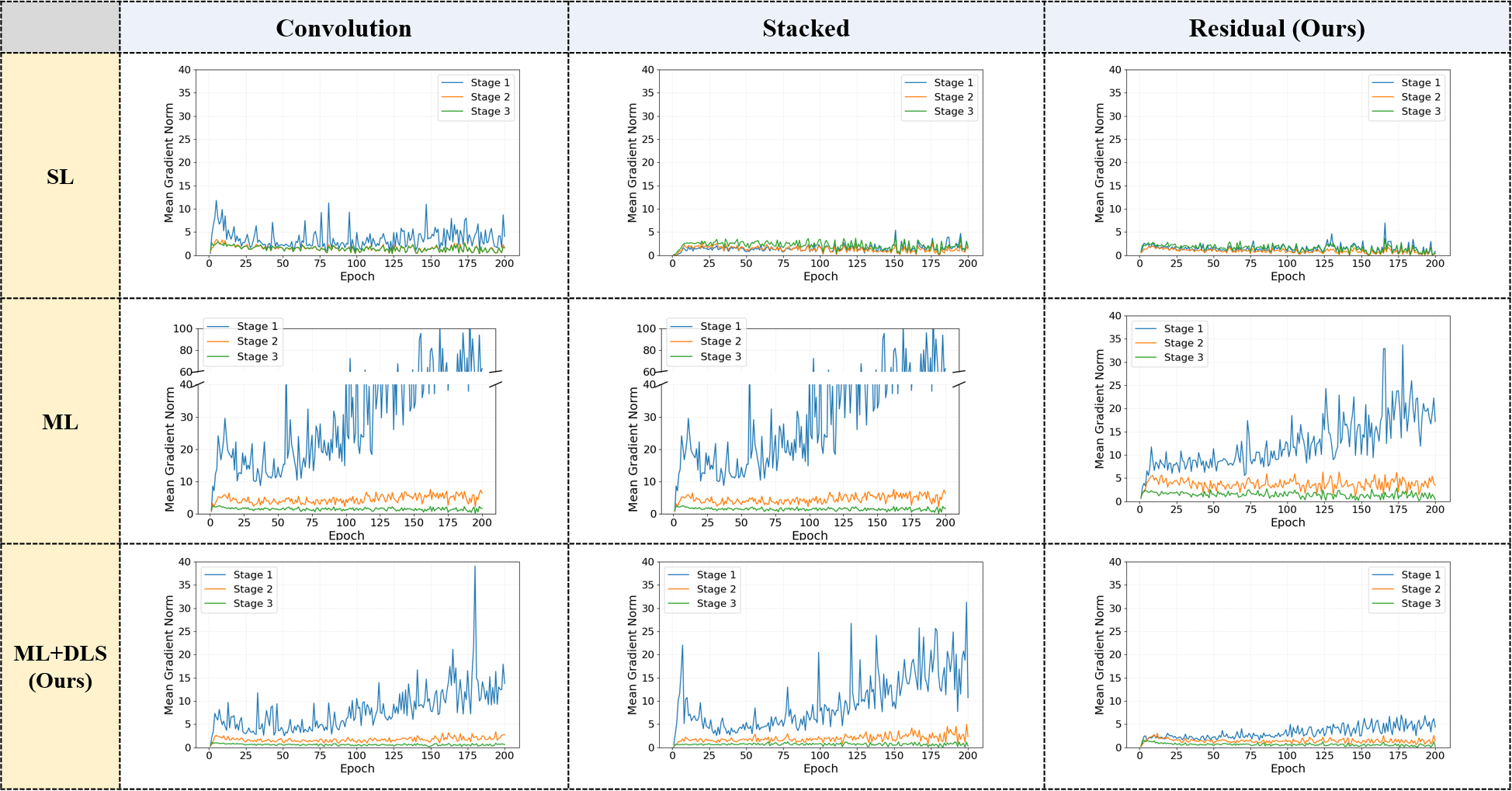} 
 \caption{Gradient norm per training epochs with ReSEViT-T model on CIFAR100. The column headers represent the token embedding variants, while the row headers represent the learning methods. SL, ML, and ML+DLS represent Single Loss, Multistage Ensemble Loss, and Multistage Ensemble Loss with Dynamic Learner Selection, respectively.}
 \label{fig:gradientnorm}
\end{figure*}

\subsection{Gradient Norm Analysis for Model Interpretation}
By comparing gradient norms across token variants and learning methods in the ReSEViT model throughout training, we examine whether certain methods yield more stable gradient magnitudes, potentially facilitating more effective gradient propagation and optimization. In Fig. \ref{fig:gradientnorm}, we observe that across most methods, gradient norms peak in Stage 1 and then decrease progressively in subsequent stages as the earlier stages are more sensitive to the input and receive stronger gradient signals. This indicates a problem when the gradients are excessively large and unstable. The ML method can further increase the gradient magnitude by accumulating gradients from multiple exit losses. To address this issue, the imbalance in gradients can be substantially reduced by the residual token and the DLS module indicating that the gradients are effectively stabilized and regulated without making them vanished. This effect is crucial particularly for small datasets, where ViT models receive less diverse supervision and are more sensitive to optimization instability. By maintaining a balanced gradient propagation across stages, residual token and DLS can improve the reliability of feature learning under limited data.

\begin{figure}[pos=!h]
\centering
\includegraphics[scale=0.6]{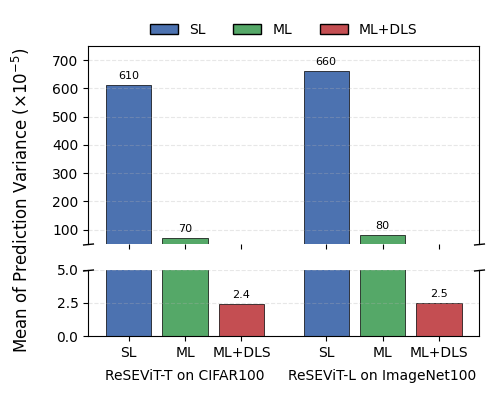}
\caption{Prediction variance of the proposed ReSEViT. SL, ML, and ML+DLS represent Single Loss, Multistage Ensemble Loss, and Multistage Ensemble Loss with Dynamic Learner Selection, respectively.}
\label{fig:variance}
\end{figure}

\subsection{Prediction Variance Analysis}
The theoretical expectation in Eq. \ref{variance} is consistent with the empirical result in Fig. \ref{fig:variance}. The suggested ML yields lower prediction variance than the SL demonstrating that averaging multiple predictions efficiently decreases variance. The DLS can significantly reduce prediction variance by averaging the predictions of multiple models and giving less influence to models with high variance. Based on the results, self-ensemble learning is helpful for ReSEViT trained on small datasets because weak inductive biases can lead to overfitting and instability in predictions. In situations with limited data, self-ensemble learning facilitates greater generalization.

\subsection{Ablation Studies}
To investigate the effectiveness of each proposed component, we conducted ablation studies. Specifically, we first verify the effects of residual token embedding. Moreover, we observe how our self-ensemble method works on the proposed model. We also investigate the effects of dynamic weight on our self-ensemble learning. Finally, we show the effects of the number of stage classifiers. 

\begin{table}[H]
\renewcommand{\arraystretch}{1.3}
\caption{Effects of Residual Token Embeddings.}
\label{table_skipconnections}
\centering
\begin{tabular}{lcll}
\hline
\bfseries Method
& \bfseries $\mathcal{L}_{WSE}$
& \bfseries CIFAR100
& \bfseries ImageNet100
\\
\hline
Conv. Token (Baseline) & \xmark & 77.28\% & 82.60\%
\\
 & \cmark & 78.62\% (+1.34) & 83.23\% (+0.63)
\\
\hline
Residual Token & \xmark & 78.65\% & 83.31\%
\\
\rowcolor{Gray}
 & \cmark & \textbf{81.05\%} (+2.40) & \textbf{85.23\%} (+1.92)
\\
\hline
\end{tabular}
\end{table}

\subsubsection{Residual Token Enhances Self-Ensemble}
Table \ref{table_skipconnections} demonstrates the importance of residual token embeddings of the ReSEViT-S and ReSEViT-L models on CIFAR100 and ImageNet100, respectively. When the convolution token is subtituted with the residual token, the model performance improves by more than 2.43\% for CIFAR100 and 2\% for ImageNet100 highlighting that improvements in local information flow can enhance the overall flow of global information. In essence, the absence of a residual token  limits the effectiveness of the self-ensemble. 

\begin{table}[ht]
\renewcommand{\arraystretch}{1.3}
\caption{Top-1 Accuracy of Self-Ensemble Method.}
\label{table_selfensemble}
\centering
\begin{tabular}{ccccc}
\hline
\bfseries Method
& \bfseries Stage 1
& \bfseries Stage 2
& \bfseries Stage 3
& \bfseries Ensemble

\\
\hline
\multicolumn{5}{c}{CIFAR100}
\\
\hline
SL & 12.22\% & 34.17\% & 78.65\% & 76.62\%
\\
ML & 65.96\% & 76.38\% & 78.22\% & 80.33\%
\\
\rowcolor{Gray}
ML+DLS & 65.01\% & 76.70\% & 78.59\% & \textbf{81.05\%}
\\
\hline
\multicolumn{5}{c}{ImageNet100}
\\
\hline
SL &  4.31\% & 13.39\% & 83.31\% & 82.52\%
\\
ML &  59.53\% & 81.05\% & 83.90\% & 84.12\%
\\
\rowcolor{Gray}
ML+DLS & 54.69\% & 81.01\% & 84.51\% & \textbf{85.23\%}
\\
\hline
\end{tabular}
\end{table}

\subsubsection{Effects of the Self-Ensemble Method}
Table \ref{table_selfensemble} shows the experimental results of our self-ensemble method on CIFAR100 with ReSEViT-S and ImageNet100 with ReSEViT-L model. Methods of SL, ML, and ML+DLS indicate single loss (baseline), multistage ensemble loss, and multistage ensemble loss with DLS, respectively. We make the following observations.

\begin{itemize}
  \item There is a clear performance gap between the earlier-stage classifiers (Stage 1 and 2) in SL and ML methods, as the ML method strengthens the supervision for all stage classifiers through error corrections against the true label.
  \item The performance of the na{\"\i}ve ensemble method (SL) drops up to more than 1.5\% due to weaker supervision at the earlier stage classifiers than that at the deepest stage classifier. In contrast, using the ML method, the ensemble of all stage classifiers improves performance.
  \item In the ML+DLS method, DLS is applied to decide which stage classifiers are highlighted most. The performance results show that the dynamic $\beta$ assigns adaptive weights to each stage classifier, thereby reducing the influence of the weaker classifiers (Stage 1 or 2) and strengthening the stronger ones. As a result, the ensemble of all classifier improves the overall prediction result.
\end{itemize}

\begin{table}[h]
\renewcommand{\arraystretch}{1.3}
\caption{Effects of Dynamic Weights (Beta).}
\label{table_beta_ap}
\centering
\begin{tabular}{c c c c c}
\hline
\bfseries Method
& \bfseries Training
& \bfseries Inference
& \bfseries CIFAR100
& \bfseries ImageNet100
\\
\hline
ML & \xmark & \xmark & 80.33\% & 84.12\% 
\\ 
ML+BT & \cmark & \xmark & 80.75\% & 84.85\%
\\ 
ML+BI & \xmark & \cmark & 80.71\% & 84.44\%
\\ 
\rowcolor{Gray}
ML+DLS & \cmark & \cmark & \textbf{81.05\%} & \textbf{85.23\%}
\\ 
\hline
\end{tabular}
\end{table}

\subsubsection{Effects of Dynamic Weights (Beta)}
To further verify the effects of DLS, we perform an ablation study with and without the dynamic weights ($\beta$) using the ReSEViT-S and ReSEViT-L models on CIFAR100 and ImageNet100, respectively, as shown in Table \ref{table_beta_ap}. ML, ML+BT, ML+BI, and ML+DLS denote Multistage Ensemble Loss, ML with $\beta$ in the loss function, ML with $\beta$ in the inference, and ML with $\beta$ in the loss function and inference, respectively. Our ML+DLS achieves the best performance, with accuracies of 81.05\% and 85.23\%. The results show that using dynamic weights in the loss function and during inference are complementary and can further boost performance.

\begin{table}[ht]
\centering
\renewcommand{\arraystretch}{1.3}
\caption{Number of Stage Classifiers.}
\label{table_numstage}
\begin{tabular}{cccccc}
    \hline
    \bfseries Num. of cls. & \bfseries Stage 1 & \bfseries Stage 2 & \bfseries Stage 3 & \bfseries CIFAR100 & \bfseries ImageNet100 \\
    
    \hline
     & \cmark & \xmark & \xmark & 68.49\% & 65.70\% \\
     1 & \xmark & \cmark & \xmark & 76.41\% & 82.67\% \\
     & \xmark & \xmark & \cmark & 78.65\% &  83.31\% \\
    \grayline
     & \cmark & \cmark & \xmark & 77.60\% & 82.63\% \\
     2 & \cmark & \xmark & \cmark & 79.66\% & 83.82\%\\
     & \xmark & \cmark & \cmark & 80.80\% & 84.81\%\\
    \grayline
    \rowcolor{Gray}
    3 & \cmark & \cmark & \cmark & \bfseries 81.05\% & \bfseries 85.23\% \\
    \hline   
\end{tabular}
\end{table}

\subsubsection{Number of Stage Classifiers}
We conduct more studies on the number of stage classifiers used for the self-ensemble based on the ReSEViT-S network. This study is helpful for understanding the effect of the applied self-ensemble on different numbers of stage classifiers. As shown in Table \ref{table_numstage}, we observe that (1) the ensemble of three classifiers achieves the best performance; (2) the deeper stage classifier obtains better performance, \textit{e.g.}, the results of the third stage classifier are better than those of the second stage classifier. Based on this study, we exploit the ensemble of three classifiers in our final design.

\begin{figure*}[pos=h]
\centering
\includegraphics[scale=0.76]{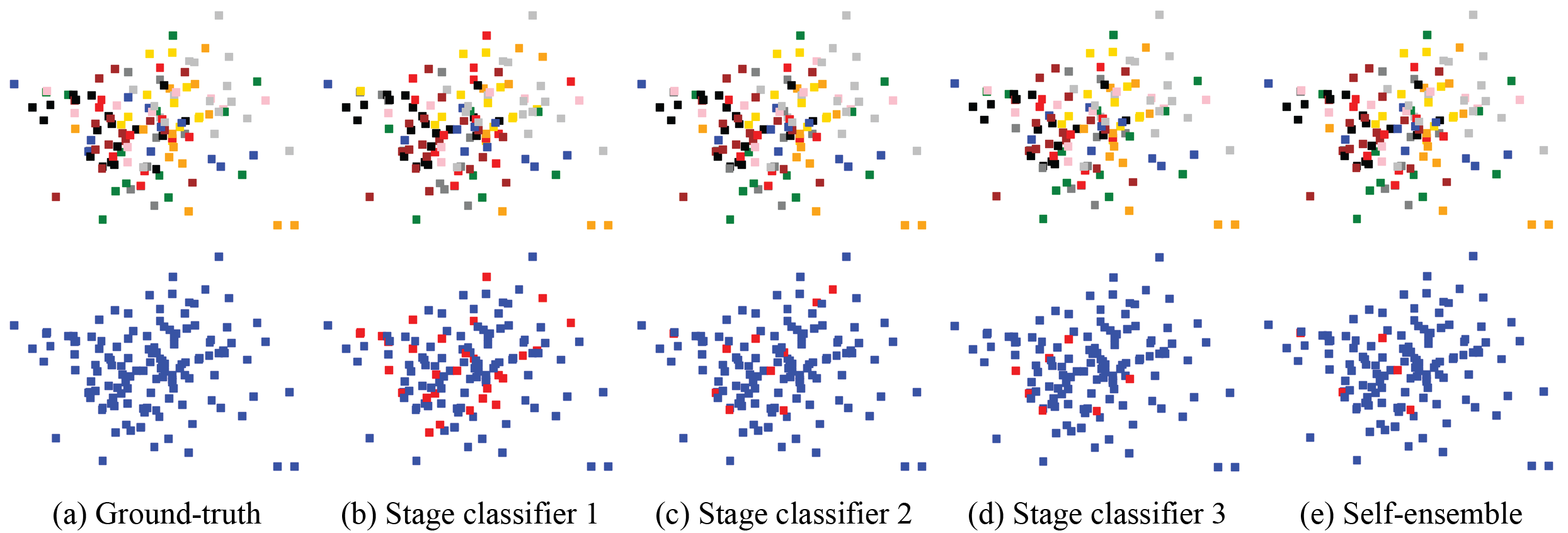}
\caption{ReSEViT-S class prediction distribution in 2-D on CIFAR10 test samples using PCA visualization. Top: classifiers class prediction distinguished by color (10 classes). Bottom: classifiers Top-1 prediction (correct predictions in blue color and incorrect predictions in red color). Best viewed in color.}
\label{fig:pca_visualize}
\end{figure*}

\subsection{Visual Interpretation}
\subsubsection{Stage Classifiers Prediction Analysis}
In this section, class prediction results on hundreds of test samples are visualized to help understand what information each classifier has learned. For a clearer presentation, we show the 2-D representation of 128 random CIFAR10 testing samples using Principle Component Analysis (PCA). We compare the results of different stage classifiers to investigate the performance of the self-ensemble method from the point of view of class prediction. Accordingly, we hope to answer the open question: \emph{``Could the self-ensemble method find a better representation and reduce the prediction error?"}

Fig.~\ref{fig:pca_visualize} shows that the deeper stage classifiers have lower prediction errors. However, the classifier in the early stage correctly predicts in some cases, while the deep-stage classifier does not. This phenomenon indicates that each stage classifier captures distinct features where the early stage still holds meaningful information that the deeper stage fails to capture. By leveraging the advantages of different classifiers, our self-ensemble method is able to find a better ensemble representation and achieve lower class prediction errors.

\begin{figure*}[pos=h]
\centering
\includegraphics[scale=1.6]{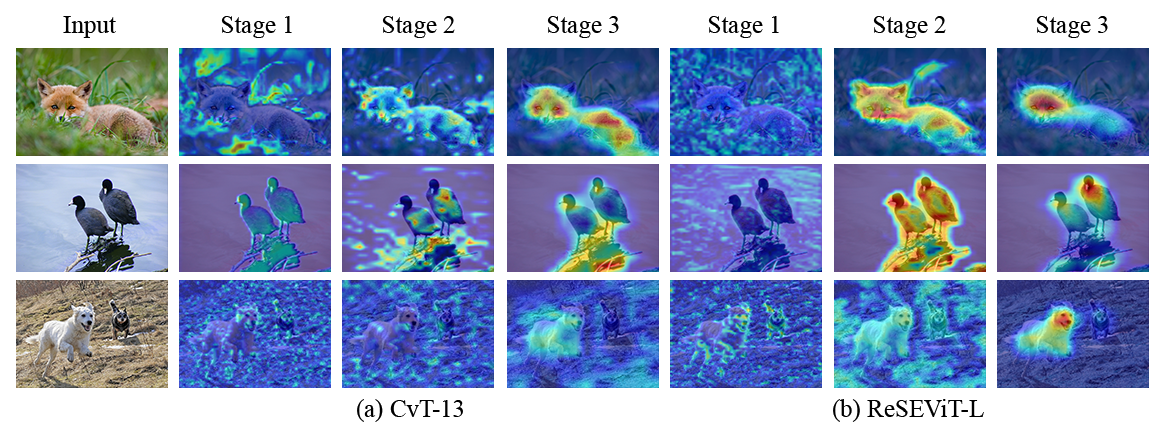}
\caption{Visualization of (a) CvT-13 (baseline) and (b) ReSEViT-L GradCAM results. Both models are trained on ImageNet100. We present visualization for each stage. Images contain single and multiple objects with various scales. The class names are Red Fox, American Coot, and Kuvasz. Best viewed in color.}
\label{fig:gradcam_visualize}
\end{figure*}

\subsubsection{Class Attention Map}
To better understand the significance of ReSEViT, we apply GradCAM \cite{gradcam} visualization of each stage, while it is generally demonstrated in the last stage only. In Fig. \ref{fig:gradcam_visualize}, two deep learning models with three-stage architecture are compared, \textit{i.e.}, CvT as a baseline, and ReSEViT. In the first stage, the activation maps have less semantic information since the network typically captures low-level features, such as edges and textures. In the second stage, ReSEViT covers single and multiple targets in the image more precisely. Such observations highlight that incorporating the local information flow from residual tokens to transformers enables ReSEViT to learn more distinctive features than CvT. In the last stage, ReSEViT is able to focus on key parts of the object more accurately with smaller activation areas. 

\section{Conclusion}
To facilitate small-scale image classification, we present ReSEViT, a multi-stage residual ViT with self-ensemble learning. It arises when ViT performs poorly on limited image data. Our self-ensemble learning alleviates the issue by leveraging information from the intermediate classifiers for global information flow. Self-ensemble learning reduces prediction variance, which can improve model generalization, especially in small-data settings. As the importance of multi-stages is not the same, we introduced DLS to optimize our ensemble performance by decision fusion weights. The DLS is proven to reduce calibration error and stabilize the gradient. To improve global information flows, we incorporate residual tokens that model stronger local structures in each model exit. The residual token can capture clearer object features that can benefit the Transformer. Unifying residual tokens and self-ensemble learning into a multi-stage ViT facilitates a more efficient computational framework reducing redundancy, prediction error, and variance. Experimental results have demonstrated that the proposed method outperforms previous methods efficiently through the multi-stage ViT design and self-ensemble learning on small image datasets.

\section*{Acknowledgments}
This work is partially supported by the National Science and Technology Council, Taiwan, under Grants: NSTC-112-2628-E-002-033-MY4 and NSTC-114-2634-F-002-004, and was financially supported in part by the Center of Data Intelligence: Technologies, Applications, and Systems, National Taiwan University (Grants: 114L900901 / 114L900902 / 114L900903), from the Featured Areas Research Center Program within the framework of the Higher Education Sprout Project by the Ministry of Education, Taiwan.
This research was supported by Culture, Sports and Tourism R{\&}D Program through the Korea Creative Content Agency grant funded by the Ministry of Culture, Sports and Tourism in 2024 (RS-2024-00398413, Contribution Rate: 15\%).


\appendix
\section{Supplementary Material}

\begin{table*}
\renewcommand{\arraystretch}{1.1}
\caption{Detailed network specification for ReSEViT-T, CvT-13, and ResNet18 on small resolution datasets (32x32). $^\dagger$ represents a modified stem for small-resolution datasets.}
\label{table_architecture_comparison_32}
\centering
\begin{adjustbox}{width=1\textwidth}
\begin{tabular}{c|l|c|l|c|l|c}
    \hline
    \bfseries Output & \bfseries Layer Name & \bfseries ReSEViT-T & \bfseries Layer Name & \bfseries CvT-13$^\dagger$ & \bfseries Layer Name & \bfseries ResNet18$^\dagger$ \\
    \hline 
    32\texttimes32 & 
    Conv. Stem & 
    3\texttimes3, 64, $s$=1 &
    \multirow{3}{*}{Conv. Token 1} & 
    \multirow{3}{*}{3\texttimes3, 64, $s$=1} &
    Conv. Stem & 
    3\texttimes3, 64, $s$=1 
    \\
    
    \cline{1-3}
    \cline{6-7}
    \multirow{2}{*}{32\texttimes32} & 
    \multirow{2}{*}{Res. Token 1} & 
    \multirow{2}{*}{
    $\Big[$
    \makecell{ 
        3\texttimes3, 64, $s$=1 \\
        3\texttimes3, 64, $s$=1
    }
    $\Big]\times2$
    }
    & 
    
    & 
    
    &
    \multirow{4}{*}{Res. Token 1} & 
    \multirow{4}{*}{$\Big[$
    \makecell{
        3\texttimes3, 64, $s$=1 \\
        3\texttimes3, 64, $s$=1
    }
    $\Big]\times2$}
    \\ 
    & & & & & &
    \\
    
    \cline{1-5}
    
    \multirow{2}{*}{32\texttimes32} & 
    \multirow{2}{*}{Transformer 1} &
    \multirow{2}{*}{
    $\Big[$
    \makecell{
        3\texttimes3, 64, $s_{k,v}$=2 \\
        $T_{1}$=32, $h_{1}$=4
    }
    $\Big]\times1$}
    & 
    \multirow{2}{*}{Transformer 1}
    &
    \multirow{2}{*}{
    $\Big[$
    \makecell{
        3\texttimes3, 64, $s_{k,v}$=2 \\
        $T_{1}$=32, $h_{1}$=1
    }
    $\Big]\times1$}
    & &
    \\
    & & & & & &
    \\
    
    \hline
    \multirow{2}{*}{16\texttimes16} & 
    \multirow{2}{*}{Res. Token 2} & 
    \multirow{2}{*}{$\Big[$
    \makecell{
        3\texttimes3, 128, $s$=2 \\
        3\texttimes3, 128, $s$=1
    }
    $\Big]\times2$}
    & 
    \multirow{2}{*}{Conv. Token 2}
    & 
    \multirow{2}{*}{3\texttimes3, 192, $s$=2} &
    \multirow{4}{*}{Res. Token 2} &
    \multirow{4}{*}{$\Big[$
    \makecell{
        3\texttimes3, 128, $s$=2 \\
        3\texttimes3, 128, $s$=1
    }
    $\Big]\times2$}
    \\
    & & & & & &
    \\
    
    \cline{1-5}
    
    \multirow{2}{*}{16\texttimes16} & 
    \multirow{2}{*}{Transformer 2} & 
    \multirow{2}{*}{$\Big[$
    \makecell{
        3\texttimes3, 128, $s_{k,v}$=2 \\
        $T_{2}$=16, $h_{2}$=4
    }
    $\Big]\times1$}
    &
    \multirow{2}{*}{Transformer 2}
    & 
    \multirow{2}{*}{$\Big[$
    \makecell{
        3\texttimes3, 192, $s_{k,v}$=2 \\
        $T_{2}$=16, $h_{2}$=3
    }
    $\Big]\times2$}
    & &
    \\
    & & & & & &
    \\
    
    \hline
    \multirow{2}{*}{8\texttimes8} & 
    \multirow{2}{*}{Res. Token 3} & 
    \multirow{2}{*}{$\Big[$
    \makecell{
        3\texttimes3, 256, $s$=2 \\
        3\texttimes3, 256, $s$=1
    }
    $\Big]\times2$}
    & 
    \multirow{2}{*}{Conv. Token 3}
    & 
    \multirow{2}{*}{3\texttimes3, 384, $s$=2}
    &
    \multirow{4}{*}{Res. Token 3}
    &
    \multirow{4}{*}{$\Big[$
    \makecell{
        3\texttimes3, 256, $s$=2 \\
        3\texttimes3, 256, $s$=1
    }
    $\Big]\times2$} 
    \\
    & & & & & &
    \\

    \cline{1-5}
    \multirow{2}{*}{8\texttimes8} & 
    \multirow{2}{*}{Transformer 3} &
    \multirow{2}{*}{$\Big[$
    \makecell{
        3\texttimes3, 256, $s_{k,v}$=2 \\
        $T_{3}$=8, $h_{3}$=4
    }
    $\Big]\times1$}
    &
    \multirow{2}{*}{Transformer 3}
    & 
    \multirow{2}{*}{$\Big[$
    \makecell{
        3\texttimes3, 384, $s_{k,v}$=2 \\
        $T_{3}$=8, $h_{3}$=6
    }
    $\Big]\times10$}
    & &
    \\
    & & & & & &
    \\
    \hline
    
    \multirow{2}{*}{4\texttimes4} & \multirow{2}{*}{Res. Token 4} & \multirow{2}{*}{-} & \multirow{2}{*}{Conv. Token 4} & \multirow{2}{*}{-} &
    \multirow{2}{*}{Res. Token 4}
    &
    \multirow{2}{*}{$\Big[$
    \makecell{
        3\texttimes3, 512, $s$=2 \\
        3\texttimes3, 512, $s$=1
    }
    $\Big]\times2$}
    
    \\
    & & & & & &
    \\
    
    \hline
    1\texttimes1 & 
    Linear &
    $100$ &
    Linear &
    $100$ &
    Linear &
    $100$ 
    \\
    
    \hline
    \multicolumn{2}{c|}{\textbf{(Params, FLOPs)}} & (3.88M, 0.56G) &  & (19.65M, 1.32G) &  & (11.22M, 0.56G)
    \\
    \hline
\end{tabular}
\end{adjustbox}
\end{table*} 

\begin{table*}
\renewcommand{\arraystretch}{1.1}
\caption{Detailed network specification for ReSEViT-T, CvT-13, and ResNet18 on a large-resolution dataset (224x224).}
\label{table_architecture_comparison_224}
\centering
\begin{tabular}{c|l|c|l|c}
    \hline
    \bfseries Output & \bfseries Layer Name & \bfseries ReSEViT-L & \bfseries Layer Name & \bfseries CvT-13 
    \\
    \hline 
    56\texttimes56 & 
    Conv. Stem & 
    7\texttimes7, 64, $s$=4 &
    \multirow{3}{*}{Conv. Token 1} & 
    \multirow{3}{*}{7\texttimes7, 64, $s$=4}
    \\
    
    \cline{1-3}
    \multirow{2}{*}{56\texttimes56} & 
    \multirow{2}{*}{Res. Token 1} & 
    \multirow{2}{*}{
    $\Big[$
    \makecell{ 
        3\texttimes3, 64, $s$=1 \\
        3\texttimes3, 64, $s$=1
    }
    $\Big]\times2$
    }
    & 
    &
    
    \\ 
    & & & &
    \\
    
    \hline
    
    \multirow{2}{*}{56\texttimes56} & 
    \multirow{2}{*}{Transformer 1} &
    \multirow{2}{*}{
    $\Big[$
    \makecell{
        3\texttimes3, 64, $s_{k,v}$=2 \\
        $T_{1}$=56, $h_{1}$=1
    }
    $\Big]\times2$}
    & 
    \multirow{2}{*}{Transformer 1}
    &
    \multirow{2}{*}{
    $\Big[$
    \makecell{
        3\texttimes3, 64, $s_{k,v}$=2 \\
        $T_{1}$=56, $h_{1}$=1
    }
    $\Big]\times1$}
    \\
    & & & & 
    \\
    
    \hline
    \multirow{2}{*}{28\texttimes28} & 
    \multirow{2}{*}{Res. Token 2} & 
    \multirow{2}{*}{$\Big[$
    \makecell{
        3\texttimes3, 192, $s$=2 \\
        3\texttimes3, 192, $s$=1
    }
    $\Big]\times2$}
    & 
    \multirow{2}{*}{Conv. Token 2}
    & 
    \multirow{2}{*}{3\texttimes3, 192, $s$=2}
    \\
    & & & & 
    \\
    
    \hline
    
    \multirow{2}{*}{28\texttimes28} & 
    \multirow{2}{*}{Transformer 2} & 
    \multirow{2}{*}{$\Big[$
    \makecell{
        3\texttimes3, 192, $s_{k,v}$=2 \\
        $T_{2}$=28, $h_{2}$=3
    }
    $\Big]\times2$}
    &
    \multirow{2}{*}{Transformer 2}
    & 
    \multirow{2}{*}{$\Big[$
    \makecell{
        3\texttimes3, 192, $s_{k,v}$=2 \\
        $T_{2}$=28, $h_{2}$=3
    }
    $\Big]\times2$}
    \\
    & & & & 
    \\
    
    \hline
    \multirow{2}{*}{14\texttimes14} & 
    \multirow{2}{*}{Res. Token 3} & 
    \multirow{2}{*}{$\Big[$
    \makecell{
        3\texttimes3, 384, $s$=2 \\
        3\texttimes3, 384, $s$=1
    }
    $\Big]\times2$}
    & 
    \multirow{2}{*}{Conv. Token 3}
    & 
    \multirow{2}{*}{3\texttimes3, 384, $s$=2} 
    \\
    & & & & 
    \\

    \hline
    \multirow{2}{*}{14\texttimes14} & 
    \multirow{2}{*}{Transformer 3} &
    \multirow{2}{*}{$\Big[$
    \makecell{
        3\texttimes3, 384, $s_{k,v}$=2 \\
        $T_{3}$=14, $h_{3}$=6
    }
    $\Big]\times3$}
    &
    \multirow{2}{*}{Transformer 3}
    & 
    \multirow{2}{*}{$\Big[$
    \makecell{
        3\texttimes3, 384, $s_{k,v}$=2 \\
        $T_{3}$=14, $h_{3}$=6
    }
    $\Big]\times10$}
    \\
    & & & &
    \\
    \hline
    1\texttimes1 & 
    Linear &
    $100$ &
    Linear &
    $100$
    \\
    
    \hline
    \multicolumn{2}{c|}{\textbf{(Params, FLOPs)}} & (12.43M, 4.08G) &  & (19.69M, 4.05G)
    \\
    \hline
\end{tabular}
\end{table*}

\subsection{Network Specification}
We show the network specification comparison between our ReSEViT-T, CvT-13$^\dagger$, and ResNet18$^\dagger$ in Table \ref{table_architecture_comparison_32}. As a multi-stage Vision Transformer, ReSEViT-T is the final form of CvT-13$^\dagger$ and ResNet18$^\dagger$. We have demonstrated previously that ResNet18$^\dagger$ performs quite well on small-resolution datasets with its efficiency through varied aspects even compared with a hybrid ViT, CvT-13$^\dagger$. Hence, this motivates us to apply it to CvT-13$^\dagger$ as a residual token embedding. We reconstruct it considering the convolution stem, network scale, network efficiency, and model stage. The experimental result shows that ReSEViT-T dominates ViTs on small-resolution image datasets. 

For a large-resolution image dataset, we design ReSEViT-L. Compared with the original CvT-13, we substitute the convolution token with residual token embeddings and adjust the transformer depth. We can see that our ReSEViT-L performance on ImageNet100 outperforms CvT-13 with less parameter size. This shows that our approach not only works well on small-resolution image datasets but also on a large-resolution dataset.

\begin{table}
\renewcommand{\arraystretch}{1.2}
\caption{Necessity of CLS Token. '/' Symbol Separates the Method without and with Self-Ensemble.}
\label{table_cls_token}
\centering
\begin{tabular}{l c c c}
\hline
\bfseries ReSEViT-S
& \bfseries [CLS]
& \bfseries GAP
& \bfseries SeqPool
\\
\hline
CIFAR100 Top-1 & 78.65$\%$ / \textbf{81.05\%} & 77.99$\%$ / 80.47$\%$ & \textbf{78.83\%} / 80.63$\%$
\\
\hline
\end{tabular}
\end{table}

\subsection{Necessity of CLS Token}
Researchers have studied the contribution of the CLS token in ViT. Concretely, CCT replaces the CLS token with sequence pooling (SeqPool), which is a learnable weighting sum method. NesT and MoCo-V3 also removed the CLS token and replaced it with Global Average Pooling (GAP). Therefore, we investigate the performance of different alternatives on ReSEViT-S. Table \ref{table_cls_token} shows that the CLS token works well in our self-ensemble model. Thus, we use the CLS tokens in different stages as our final design.

\begin{figure}[pos=t]
\centering
\includegraphics[scale=0.545]{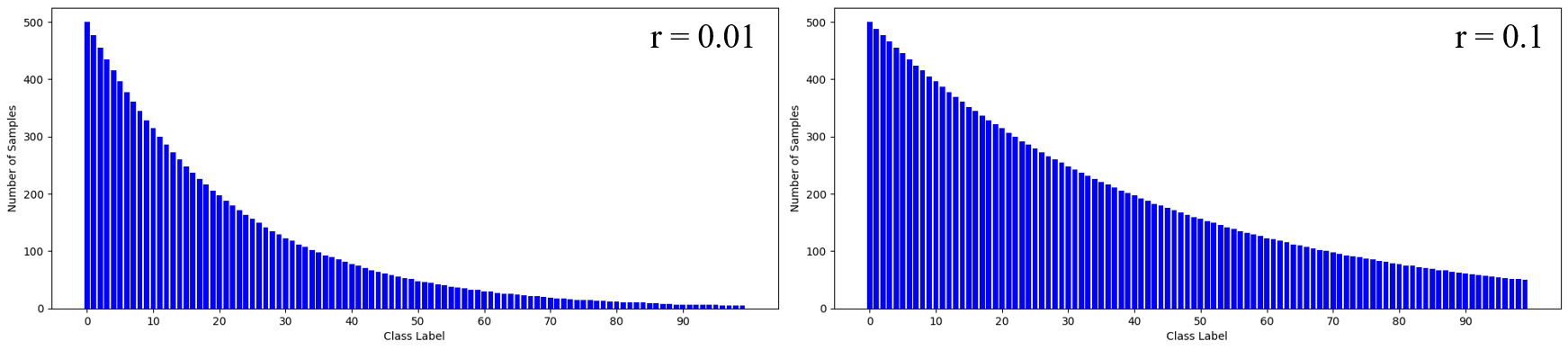}
\caption{Class Imbalance Distribution across various imbalance ratio ($r$).}
\label{fig:imbalancedistribution}
\end{figure}

\begin{table}
\renewcommand{\arraystretch}{1.3}
\caption{Comparison with other models under class imbalance. $^\dagger$ refers to a modified stem. We report accuracy and ECE.}
\label{table_imbalance}
\centering
\begin{tabular}{c|cc|cc|cc}
\multirow{3}{*}{\bfseries Method} 
& \multicolumn{6}{c}{\bfseries CIFAR-100}
\\
\cline{2-7}
& \multicolumn{2}{c|}{\bfseries ($r=$ 0.01)}
& \multicolumn{2}{c|}{\bfseries ($r=$ 0.1)}
& \multicolumn{2}{c}{\bfseries ($r=$ 1.0)}
\\
\cline{2-7}
 & \bfseries Top-1 $\uparrow$ & \bfseries ECE $\downarrow$ & \bfseries Top-1 $\uparrow$ & \bfseries ECE $\downarrow$ & \bfseries Top-1 $\uparrow$ & \bfseries ECE $\downarrow$
\\
\hline
ResNet18$^\dagger$ \cite{resnet} & 36.82$\%$ & 0.145 & 56.71$\%$ & 0.144 & 75.41$\%$ & 0.158
\\
ResNet34$^\dagger$ \cite{resnet} & 38.67$\%$ & 0.168 & 58.36$\%$ & 0.141 & 77.08$\%$ & 0.126
\\
ConvNeXt-T$^\dagger$ \cite{convnext} & 34.45$\%$ & 0.221 & 54.96$\%$ & 0.199 & 74.02$\%$ & 0.113
\\
CvT-13$^\dagger$ \cite{cvt} & 38.45$\%$ & 0.221 & 56.77$\%$ & 0.151 & 73.56$\%$ & 0.115
\\
CCT-7/3$\times$1 \cite{cct} & 39.58$\%$ & 0.161 & 58.72$\%$ & 0.134 &  76.67$\%$ & 0.132
\\
NesT-T \cite{nest} & 41.89$\%$ & 0.154 & 60.90$\%$ & 0.143 & 78.69$\%$ & 0.147
\\
\hline
ReSEViT-T (w/o $\mathcal{L}_{WSE}$) & 42.54$\%$ & 0.164 & 61.65$\%$ & 0.129 & 77.65$\%$ & 0.123
\\
ReSEViT-T (w/o DLS) & 42.60$\%$ & 0.441 & 63.36$\%$ & 0.272 & 78.60$\%$ & 0.160
\\
\rowcolor{Gray}
\textbf{ReSEViT-T (Ours)} & \textbf{43.85\%} & 0.147 & \textbf{64.10\%} & 0.135 & \textbf{79.49\%} & 0.146
\\
\hline
\end{tabular}
\end{table}

\subsection{Training under Class Imbalance}
We evaluate ReSEViT in a class imbalance scenario, as many small datasets often face this issue. We adopt an imbalance setup based on \cite{imbalance}, which exponentially alters the data distribution across classes with class-imbalanced ratios $r$ on the CIFAR-100 training set. The $r$ indicates the proportion of samples in the rarest class and the most frequent class. In Fig. \ref{fig:imbalancedistribution}, we present the distribution of class imbalance on the CIFAR100 training set. A dataset with an imbalance ratio ($r$) of 0.1 suggests that the majority class outnumbers the minority class by a factor of ten. On the other hand, $r$ of 0.01 means that the majority class has a hundred times more samples than the minority class. In Table \ref{table_imbalance}, we report two metrics, i.e., top-1 accuracy and expected calibration error (ECE), to observe the effect on class-imbalance settings and calibration since the accurate estimation of model calibration is crucial for the safe application of Neural Networks \cite{calibration}. 

\subsubsection{Performance Results on Class Imbalance}
We observe that ReSEViT-T outperforms the other model architectures in Top-1 accuracy across different $r$. Additionally, the proposed self-ensemble and DLS modules work well even under different $r$. As the dataset becomes highly imbalanced, the performance degradation of our modules remains relatively stable. 

\subsubsection{ECE Analysis}
ECE is also evaluated to estimate the model calibration under the class imbalance task. We observe that the majority of models have better calibration as the dataset is highly balanced. Furthermore, the incorporation of DLS modules leads to a substantial improvement in calibration relative to the performance of a standalone self-ensemble. This implies that the dynamic weighting of each classifier is very effective in terms of accurate calibration.

\begin{table*}[ht]
\renewcommand{\arraystretch}{1.3}
\caption{A roadmap to ReSEViT on CIFAR100. Top-1 accuracy, parameter size, floating point operations per second (FLOPs), and inference throughput (images/s) are reported. $^\star$ represents the use of convolution tokens with wider channel sizes for comparison.}
\label{table_roadmap}
\centering
\resizebox{\textwidth}{!}{
\begin{tabular}{r|ccccccccc}
Roadmap & CvT-13 & $\rightarrow$ & patchify "stem" & $\rightarrow$ & stage ratio & $\rightarrow$ & res. token (ResNet) & $\rightarrow$ & ReSEViT-T
\\
Note & (large resolution setting) & & (small resolution setting) & & (base form) & & (wide conv.$^\star$ / residual block) & & (final form)
\\
Accuracy ($\uparrow$) & 58.48\% & & 73.56\% & & 67.69\% & & 75.86\%$^\star$ / 77.65\% & & \textbf{79.49\%}
\\
Throughput ($\uparrow$) & 2781 & & 2560 & & 7130 & &  4254$^\star$ / 4841 & & 4832
\\
Params. ($\downarrow$) & 19.66M & & 19.65M & & 1.57M & &  5.85M$^\star$ / 3.86M & & 3.88M
\\
FLOPs ($\downarrow$) & 0.10G & & 1.32G & & 0.22G & &  0.84G$^\star$ / 0.56G & & 0.56G
\\
\hline
Roadmap & ResNet18 & $\rightarrow$ & patchify "stem" & $\rightarrow$ & w/o 4$^{th}$ stage & $\rightarrow$ & transformer & $\rightarrow$ & ReSEViT-T
\\
Note & (large resolution setting) & & (small resolution setting) & & (base form) & & (conv. transformer) & & (final form)
\\
Accuracy ($\uparrow$) & 57.15\% & & 75.41\% & & 72.85\% & & 77.65\% & & \textbf{79.49\%}
\\
Throughput ($\uparrow$) & 15544 & & 10351 & & 14292 & & 4841 & & 4832
\\
Params. ($\downarrow$) & 11.23M & & 11.22M & & 2.80M & & 3.86M & & 3.88M
\\
FLOPs ($\downarrow$) & 0.04G & & 0.56G & & 0.42G & & 0.56G & & 0.56G
\end{tabular}}
\end{table*}

\subsection{Roadmap Design for Small Dataset}
In this section, we present a detailed projection for multi-stage ViT design from CvT-13 to ReSEViT-T that is used for small-data image classification in Table \ref{table_roadmap}. As shown, the design retains its final shape from ResNet18 to ReSEViT-T, bridging the gap between the ResNet and the Transformer. Moreover, it has been acknowledged that when the deep learning model's performance improves, the model size will also be larger. In this work, we optimize our multi-stage ViT design to achieve the ideal equilibrium between accuracy and efficiency. 
\subsubsection{Patchify} Originally, the CvT-13 \cite{cvt} was designed for large resolution images and had an original convolutional stem of 7$\times$7 kernel size with 4 strides to reduce the input from 224$\times$224 to 56$\times$56. For the small-resolution input design (32$\times$32), we replace it with a 3$\times$3 kernel size with 1 stride to increase the effective receptive fields. As a result, the accuracy is drastically improved by 15.08\% showing that stem design is important for a small-resolution input size.

\subsubsection{Stage Ratio} Afterward, the stage ratio between convolution channel size and transformer depth into the base form is shrunk as the parameter size and the FLOPs are excessive for small-scale datasets compared with CCT-7/3$\times$1 and Nest-T (around 3-6M parameter size). Specifically, the convolution channel size is reduced from [64,192,384] to [64,128,256], and the transformer depth is reduced from [1,2,10] to [1,1,1]. The base network becomes super fast and light, but the performance reduces by 5.87\%. 

\subsubsection{Local Information Flow} To compensate for the ideal computational budget, we substitute the convolution token with the residual token. As the 4$^{th}$ stage of ResNet18 with 512 channel size is redundant for small-scale design (significantly increases the parameters to 11.22M), we reconstruct it with the same design principle of CvT-13 to the three-stage model. Notably, the trade-off between accuracy and efficiency is decent. Specifically, the parameter size and FLOPs are comparable with other ViTs \cite{cct, nest}, and the accuracy increases by 9.96\%. Afterward, residual tokens and convolution tokens with wider channel sizes from [64, 128, 256] to [128, 256, 512] are also compared. We observe that the residual token outperforms wider convolution token as the Transformer is able to process stronger local structures more effectively.

\subsubsection{Global Information Flow} Finally, self-ensemble learning leverages the multi-level predictions from different classifiers. By applying the self-ensemble, the performance improves by 1.84\% while maintaining a similar model size and speed. Comparing upstream and downstream, ReSEViT-T performs 5.93\% better than CvT-13, with faster inference time and a smaller model size. This demonstrates the advantages of our method for improving CvT-13 in an efficient manner for small-scale data.

\begin{figure*}[pos=t]
 \centering
 \includegraphics[scale=0.269]{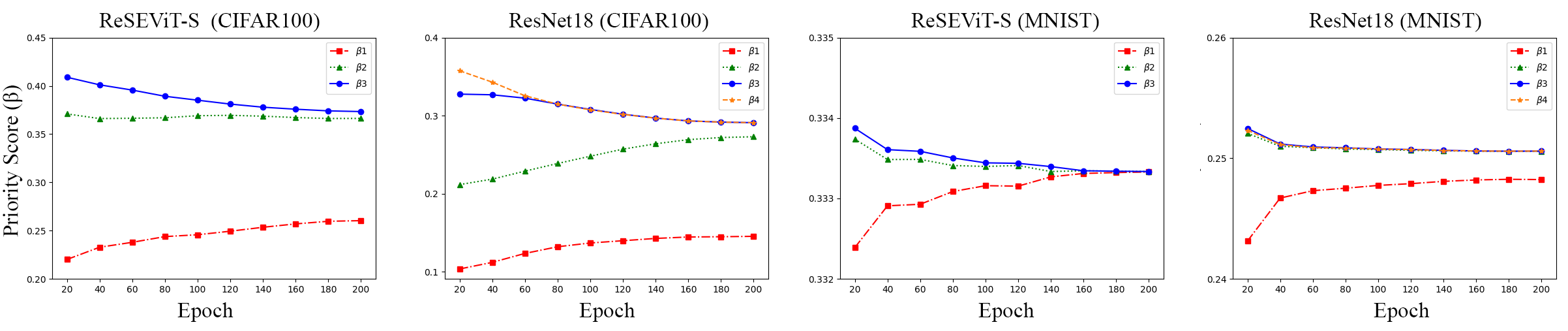} 
 \caption{Priority scores of ReSEViT-S and ResNet18 with different numbers of epochs on CIFAR100 and MNIST datasets.}
 \label{fig:betascore}
\end{figure*}

\subsection{Priority Score of Dynamic Analysis}
We show the priority scores $\beta_{n}$ of $n$-th stage classifier with different numbers of epochs for ReSEViT-S and ResNet18 on both the complex dataset (CIFAR100) and simple dataset (MNIST) in Fig. \ref{fig:betascore}. We make the following observations:
\begin{itemize}
  \item Overall, deeper stage classifiers are given a larger initial value of $\beta$. This indicates that the earlier stage classifiers do not perform as well as the deeper ones in the early epochs. 
  \item As the number of epochs increases, the priority scores dynamically change. We show that the $\beta$s become more steady when each stage classifier reaches its maximum accuracy in the later epoch. 
  \item The gap between $\beta_{1}$ and the others is smaller in the last epoch on the MNIST dataset. This suggests that different stage classifiers are critical for ensembles, although the deeper classifiers outperform the earlier ones.
\end{itemize}

\begin{table}[ht]
\renewcommand{\arraystretch}{1.1}
\caption{Our Self-Ensemble Method on Various Deep Learning Models. `SE-' Variants Utilize Self-Ensemble. $^\dagger$ refers to the modified stem. $S$ denotes the number of stages.}
\label{table_selfensemblevarious}
\centering
\begin{tabular}{lcccc}
\hline
\bfseries Model
& \bfseries $S$
& \bfseries Top-1 (\%)
& \bfseries Params. (M) $\downarrow$
& \bfseries FLOPs. (G) $\downarrow$
\\
\hline
SqueezeNet \cite{squeezenet} & \multirow{2}{*}{4} & 58.08 & 0.78 & 0.05
\\
SE-SqueezeNet & & 66.48 (+8.40) & 0.90 (+0.12) & 0.07 (+0.02)
\\
\grayline
ResNet18$^\dagger$ \cite{resnet} & \multirow{2}{*}{4} & 75.41 & 11.22 & 0.56
\\
SE-ResNet18$^\dagger$ & & 77.05 (+1.64) & 11.27 (+0.05) & 0.56
\\
\grayline
ResNet34$^\dagger$ \cite{resnet} & \multirow{2}{*}{4} & 77.08 & 21.33 & 1.16
\\
SE-ResNet34$^\dagger$ & & 78.43 (+1.35) & 21.37 (+0.04) & 1.16
\\
\grayline
PreAct-ResNet1001 \cite{preactresnet} & \multirow{2}{*}{4} & 78.48 & 10.35 & 1.53
\\
SE-PreAct-ResNet1001 & & 79.57 (+1.09) & 10.37 (+0.02) & 1.53
\\
\grayline
CvT-13$^\dagger$ \cite{cvt} & \multirow{2}{*}{3} & 73.56 & 19.65 & 1.32
\\
SE-CvT-13$^\dagger$ & & 76.52 (+2.96) & 19.68 (+0.03) & 1.32
\\
\grayline
ConvNeXt-T$^\dagger$ \cite{convnext} & \multirow{2}{*}{4} & 74.02 & 27.94 & 1.45
\\
SE-ConvNeXt-T$^\dagger$ & & 77.34 (+3.32) & 27.96 (+0.02) & 1.45
\\
\grayline
PoolFormer-S36$^\dagger$ \cite{poolformer} & \multirow{2}{*}{4} & 76.44 & 30.39 & 1.62
\\
SE-PoolFormer-S36$^\dagger$ & &  79.79 (+3.35) & 30.45 (+0.06) & 1.62
\\
\grayline
NesT-T \cite{nest} & \multirow{2}{*}{3} & 78.69 & 6.20 & 2.25
\\
SE-NesT-T & & 79.85 (+1.16) & 6.23 (+0.03) & 2.25
\\
\grayline
Conv2Former-N$^\dagger$ \cite{conv2former} & \multirow{2}{*}{4} & 77.19 & 13.83 & 0.73
\\
SE-Conv2Former-N$^\dagger$ &  &  78.34 (+1.15) & 13.96 (+0.13) & 0.76 (+0.03)
\\
\hline
\rowcolor{Gray}
Average & & \textbf{+2.71\%} & +0.06M & $<$0.01G
\\ 
\hline
\end{tabular}
\end{table}

\subsection{Compatibility of Self-Ensemble Method}
Since the proposed self-ensemble method is model-agnostic, the proposed self-ensemble method can be applied to other deep learning models, including CNNs and vision transformer models. Table \ref{table_selfensemblevarious}~shows the results with different architectures on CIFAR100. We observe that our self-ensemble method consistently improves the accuracy of different deep-learning models. On average, the accuracy improvement is 2.71\% with the additional parameter of 0.06M, while the additional FLOPs are less than 0.01G. This suggests that the self-ensemble method efficiently and consistently boosts various model performances.

\begin{figure*}[pos=!h]
\centering
\includegraphics[scale=0.93]{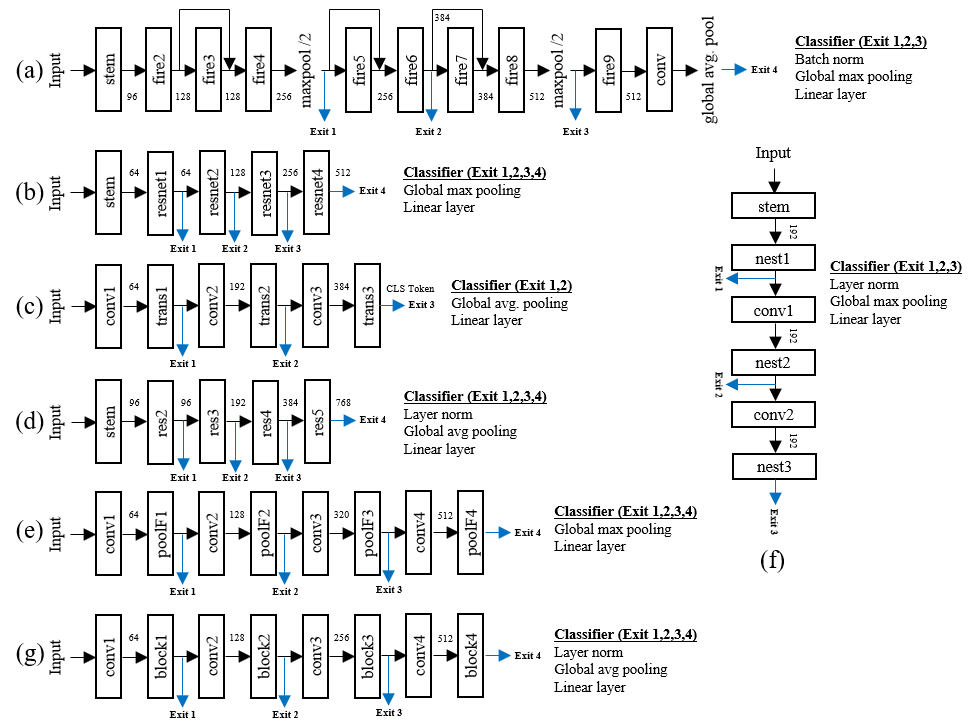}
\caption{Details of various multi-exit model architectures. (a) SE-SqueezeNet; (b) Applies to SE-ResNets and SE-PreAct-ResNet1001; (c) SE-CvT-13$^\dagger$; (d) SE-ConvNeXt-T$^\dagger$; (e) SE-PoolFormer-S36$^\dagger$; (f) SE-NesT-T; (g) SE-Conv2Former-N$^\dagger$.}
\label{fig:detail_multiexit}
\end{figure*}

We present detailed information of Table \ref{table_selfensemblevarious} on the self-ensemble application across various deep learning models in Fig. \ref{fig:detail_multiexit}. Each model has a varying number of stages, reflecting different levels of feature extraction and decision-making processes. Additionally, the classifiers within these models differ in design.

\begin{table}[!ht]
\renewcommand{\arraystretch}{1.3}
\caption{Performance of ViT variants on ImageNet100 with timm library.}
\label{table_trainvit_normal}
\centering
\begin{tabular}{c|cc|cc}
\hline
\bfseries \multirow{3}{*}{Method} & \multicolumn{4}{c}{\bfseries Training set}
\\
\cline{2-5}
 & \multicolumn{2}{c|}{\bfseries 10$\%$}  &  \multicolumn{2}{c}{\bfseries 100$\%$}
\\
& \bfseries Top-1
& \bfseries Top-5
& \bfseries Top-1
& \bfseries Top-5
\\
\hline
ViT-B & 33.68$\%$ & 61.04$\%$ & 83.74$\%$ & 95.42$\%$ 
\\
CvT-13 & 56.66$\%$ & 81.28$\%$ & 89.70$\%$ & 97.72$\%$ 
\\
\rowcolor{Gray}
\textbf{ReSEViT-L} & \textbf{64.20$\%$} & \textbf{86.46$\%$} & \textbf{90.02$\%$} & \textbf{98.18$\%$} 
\\
\hline
\end{tabular}
\end{table}

\subsection{ViT Variants Performance with Timm Library}
To assess the generality under common training configurations for ImageNet-1K, the models are trained from scratch for 300 epochs on the ImageNet100 dataset using the timm library, which is commonly used for ImageNet-1K experiments. Specifically, we use the same training settings as CvT-13, with the same data augmentation. Unless otherwise stated, all models are trained with a batch size of 64. We can see that ViT-B suffers from data-hungry issues. As the training set drops to 10$\%$, our ReSEViT-L significantly outperforms the other. Based on the results, our approach is highly advantageous in situations with limited available data.

\begin{figure}[pos=!h]
\centering
\includegraphics[scale=0.5]{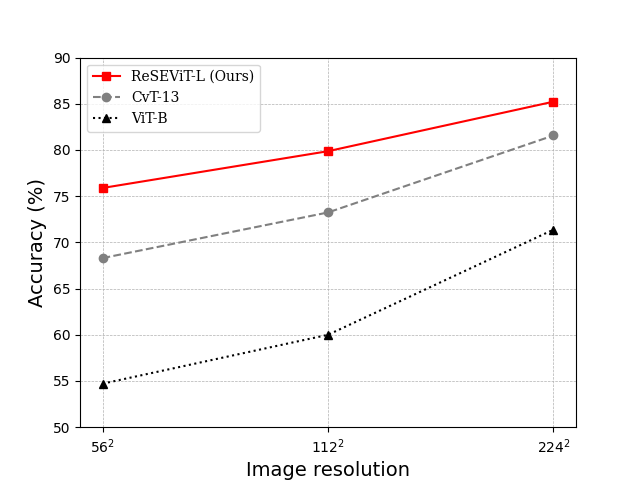}
\caption{Comparison of ViTs against image resolution on ImageNet100.}
\label{fig:imageresolution}
\end{figure}

\begin{table}[H]
\renewcommand{\arraystretch}{1.5}
\caption{Comparison between image resolution and model computation across various stem designs on ImageNet100.}
\label{table_computation}
\centering
\begin{tabular}{c|ccc|ccc|ccc}
\multirow{2}{*}{\bfseries Method}
& \multicolumn{3}{c|}{56$\times$56}
& \multicolumn{3}{c|}{112$\times$112}
& \multicolumn{3}{c}{224$\times$224}
\\
\cline{2-10}
& \bfseries Param.
& \bfseries FLOPs
& \bfseries FPS
& \bfseries Param.
& \bfseries FLOPs
& \bfseries FPS
& \bfseries Param.
& \bfseries FLOPs
& \bfseries FPS
\\
\hline
ViT-B \cite{vit} & 85.29M & 16.74G & 595 & 85.41M & 16.76G & 592 & 85.85M & 16.85G & 605
\\
CvT-13 \cite{cvt} & 19.65M & 4.02G & 990 & 19.65M & 4.02G & 988 & 19.69M & 4.05G & 1034
\\
ReSeViT-L & 12.42M & 4.06G & 866 & 12.42M & 4.06G & 870 & 12.43M & 4.08G & 890
\\
\end{tabular}
\end{table}

\subsection{Sensitivity against Image Resolution}
 We evaluate the sensitivity of our approach compared with other ViT models against image resolution in Fig. \ref{fig:imageresolution}. For a fair comparison, we start by changing the stem of each model. For ViT-B, we change the patch size of [16$^2$, 8$^2$, 4$^2$] for the input size of [224$^2$, 112$^2$, 56$^2$], respectively. For CvT-13 and ReSEViT-L, we change the convolution stem of [(7x7, s=4, p=2), (3x3, s=2, p=1), (3x3, s=1, p=1)] for the input size of [224$^2$, 112$^2$, 56$^2$], respectively. In Table \ref{table_computation}, we observe that the computational cost exhibits minimal differences when the patch size and the convolution stem are set to accommodate different image scales. The ViT models are less sensitive to image resolution within the same effective receptive fields compared with the training data size. Hence, the data-hungry problem is surely caused by the data size concluding that more training data is more important than increasing resolution for ViTs.

\begin{table}[H]
\renewcommand{\arraystretch}{1.3}
\caption{Training config.}
\label{table_config}
\centering
\begin{tabular}{l|ccc}
\bfseries Dataset & CIFAR (32$\times$32), Tiny-ImageNet (64$\times$64),  & ImageNet100 & ISIC 2018
\\
 & MedMNIST (28$\times$28) &  & 
\\
\hline
optimizer & AdamW & AdamW & AdamW
\\
init lr & 5e-4 & 5e-4 & 5e-4
\\
batch size & 64 & 64 & 64
\\
weight decay & 3e-2 & 3e-2 & 3e-2
\\
momentum & (0.9, 0.999) & (0.9, 0.999) & (0.9, 0.999)
\\
training epochs & 200 & 100 & 100
\\
lr decay & cosine & cosine & cosine
\\
warmup epochs & 10 & 10 & 10
\\
label smoothing & 0.1 & 0.1 & weighted (0.1)
\\
\hline
training & random crop (p=4) & random resize crop (224) & resize (224)
\\
augmentation & horizontal flipping & horizontal flipping & horizontal flipping
\\
& cifar auto-augment & imagenet auto-augment & vertical flipping
\\
& & & random rotation (180)
\\
& & & color jitter (0.2)
\\
& & & gaussian blur (3)
\\
\hline
testing & resize & resize (256) & resize (224)
\\
augmentation &  & center crop (224) &
\\
\end{tabular}
\end{table}

\subsection{Training Settings}
We provide detailed training config in Table \ref{table_config}. Our model is trained from scratch on two input resolutions (32$^2$ and 224$^2$) with one GPU. The GPU used is an Nvidia RTX 3090 with 24GB of memory. Our code is implemented by PyTorch. For the ISIC 2018 dataset, the color
jitter includes brightness, contrast, hue, and saturation.

\bibliographystyle{cas-model2-names}

\bibliography{IVC/cas-refs}


\vskip3pc

\end{document}